\documentclass[conference]{IEEEtran}
\IEEEoverridecommandlockouts
\usepackage{amsmath,amssymb,amsfonts}
\usepackage{graphicx}
\usepackage{url}
\usepackage{microtype}
\usepackage{cite}
\usepackage{textcomp}
\usepackage{hyperref}
\def\BibTeX{{\rm B\kern-.05em{\sc i\kern-.025em b}\kern-.08em
    T\kern-.1667em\lower.7ex\hbox{E}\kern-.125emX}}

\usepackage{algorithm}
\usepackage{algorithmicx}
\usepackage{algpseudocode}
\makeatletter
\newcommand{\multiline}[1]{%
  \begin{tabularx}{\dimexpr\linewidth-\ALG@thistlm}[t]{@{}X@{}}
    #1
  \end{tabularx}
}
\makeatother

\makeatletter
\newcommand{\linebreakand}{%
  \end{@IEEEauthorhalign}
  \hfill\mbox{}\par
  \mbox{}\hfill\begin{@IEEEauthorhalign}
}
\makeatother

\usepackage{booktabs}
\usepackage{tabularx}
\newcolumntype{Y}{>{\centering\arraybackslash}X}
\usepackage{multirow}

\newcommand{\comment}[1]{}
\usepackage{xcolor}

\newcommand{\myparagraph}[1]{\vspace{2ex}\noindent\textbf{#1}}

\title{Scaling Survival Analysis in Healthcare with Federated Survival Forests: A Comparative Study on Heart Failure and Breast Cancer Genomics}

\author{
    \IEEEauthorblockN{Alberto Archetti}
    \IEEEauthorblockA{\textit{DEIB}\\
    \textit{Politecnico di Milano}\\
    Milan, Italy\\
    alberto.archetti@polito.it}
\and
    \IEEEauthorblockN{Francesca Ieva}
    \IEEEauthorblockA{\textit{Department of Mathematics}\\
    \textit{Politecnico di Milano}\\
    Milan, Italy\\
    francesca.ieva@polimi.it}
\and
    \IEEEauthorblockN{Matteo Matteucci}
    \IEEEauthorblockA{\textit{DEIB}\\
    \textit{Politecnico di Milano}\\
    Milan, Italy\\
    matteo.matteucci@polimi.it}
}

\begin{document}

\maketitle

\begin{abstract}
Survival analysis is a fundamental tool in medicine, modeling the time until an event of interest occurs in a population. However, in real-world applications, survival data are often incomplete, censored, distributed, and confidential, especially in healthcare settings where privacy is critical. The scarcity of data can severely limit the scalability of survival models to distributed applications that rely on large data pools.
Federated learning is a promising technique that enables machine learning models to be trained on multiple datasets without compromising user privacy, making it particularly well-suited for addressing the challenges of survival data and large-scale survival applications. Despite significant developments in federated learning for classification and regression, many directions remain unexplored in the context of survival analysis.
In this work, we propose an extension of the Federated Survival Forest algorithm, called FedSurF++. This federated ensemble method constructs random survival forests in heterogeneous federations. Specifically, we investigate several new tree sampling methods from client forests and compare the results with state-of-the-art survival models based on neural networks. The key advantage of FedSurF++ is its ability to achieve comparable performance to existing methods while requiring only a single communication round to complete.
The extensive empirical investigation results in a significant improvement from the algorithmic and privacy preservation perspectives, making the original FedSurF algorithm more efficient, robust, and private.
We also present results on two real-world datasets -- a heart failure dataset from the Lombardy HFData project and Fed-TCGA-BRCA from the Falmby suite -- demonstrating the success of FedSurF++ in real-world healthcare studies. Our results underscore the potential of FedSurF++ to improve the scalability and effectiveness of survival analysis in distributed settings while preserving user privacy.


\end{abstract}

\section{Introduction}
\label{sec:introduction}

Survival analysis, or time-to-event analysis, is a branch of statistical machine learning that models the time until an event occurs in a population~\cite{klein2003survival}. It is an essential tool for clinical trials, used to compare the survival rates of different treatments or groups of patients and to study the factors that influence disease onset or progression~\cite{wang2019machine}. The goal of a survival model is to construct a survival function for a given subject in the population. The survival function
\begin{equation}
S(t) = P(T > t)
\end{equation}
represents the probability that the subject will not experience, or \emph{survive}, a given event by time $t$. Survival models use data to estimate the survival function, however, most healthcare applications involve data that are distributed across multiple devices, scarce, and confidential~\cite{andreux2020federated,rieke2020future}. Additionally, some data may have incomplete information about the subjects' survival time, a phenomenon called \emph{censoring}. For example, when studying the survival rates of patients with a certain disease, some patients may drop out of the trial or be alive at the end of the trial, making their true survival time unknown. Censoring is a common challenge in survival analysis because it can bias results and reduce the statistical power of the analysis. Increasing the number of data samples for training could help, but this is often not feasible due to difficulties in data collection and confidentiality constraints.

To overcome these limitations, Federated Learning (FL)~\cite{li2020federated,kairouz2021advances} has emerged as a promising technique to improve the success of survival applications in large-scale real-world scenarios. FL allows multiple parties with private data sets to collaboratively train a machine learning model without sharing private data information. Private data remain on the storage device, ensuring confidentiality for agents in the federation. Federated models have better generalization performance than local models because they can leverage a large and representative data pool. FL has great potential in scenarios with small local privacy-protected datasets, such as clinics and hospitals, where each data sample is valuable and private.

Federated survival analysis aims to develop techniques for applying survival models in federated settings. Several survival studies have used federated learning to analyze clinical data from different domains, such as cancer genomics~\cite{andreux2020federated,federated2022lu,terrail2022flamby}, stroke detection~\cite{learning2020duan}, and COVID-19 survival~\cite{wang2022survmaximin}. These studies mostly use either non-parametric methods, such as Kaplan-Meier estimators~\cite{truly2021froelicher}, or semi-parametric Cox models~\cite{lu2015webdisco,andreux2020federated,learning2020duan,dssurvival2022banerjee,verticox2022dai,larynx2022hansen,accurate2022kamphorst,federated2022masciocchi,wang2022survmaximin,imakura2023dccox,zhang2023federated}. However, the Cox model is based on the proportional hazard assumption, which may not be true in large federated datasets. In addition, Cox models have a linear relationship between covariates and survival ratios across subjects, which facilitates their interpretation but limits their modeling power. Some recent works have extended federated survival analysis to non-linear models based on neural networks~\cite{federated2022lu,rahimian2022practical,rahman2022fedpseudo}. This is an emerging line of research that enables survival analysis on large distributed datasets. However, most of the works use survival datasets only for benchmarking and do not address clinically relevant questions or conduct clinical trials~\cite{chowdhury2021review}.

This paper presents an extension of the Federated Survival Forest (FedSurF) algorithm~\cite{archetti2023federated}, called FedSurF++, that applies Random Survival Forests (RSFs)~\cite{ishwaran2008random} in a federated setting. RSFs are tree-based models that can handle censored data, missing values, and categorical variables. They also have lower computational complexity and higher interpretability than neural networks due to their tree-based nature. FedSurF++ exploits the advantages of RSFs and adapts them to the federated environment, where data are distributed across multiple clients and cannot be shared. The key idea of FedSurF++ is to train a local random survival forest on each client's private data and then build a federated ensemble of trees on the central server. The server selects the local client trees with a sampling method proportional to their performance metrics computed on a local validation split. This way, FedSurF++ can train a global RSF model from local RSFs with only one round of communication, reducing communication overhead and latency compared to iterative federated learning algorithms. As shown in~\cite{archetti2023federated}, the final global model consists of the best-performing trees, increasing the model's expressiveness in heterogeneous scenarios.

With this work, our contribution is twofold. First, we extend and investigate new sampling techniques for tree selection based on standard survival metrics. The results show how including a metric evaluation step to select the best trees is generally more powerful than sampling trees at random. Also, while metric evaluation increases model performance in heterogeneous settings, the specific tree-sampling metric does not affect in a statistically significant way the effectiveness of the final model. Therefore, the simplest evaluation metric, such as concordance, is sufficient to obtain the best final model. Second, we apply FedSurF++ to two real-world cases. The first is an administrative study concerning hospitalizations of patients experiencing heart failure~\cite{Mazzali2016MethodologicalIO}. The dataset comes from the Lombardy HFData research project and is composed of 895 samples with 32 covariates split across 23 medical institutes. The second comes from the dataset suite called (Flamby)~\cite{terrail2022flamby} and collects 38 finary features for each of the 1088 patients suffering from breast invasive carcinoma. Results on both datasets show how FedSurF++ remains competitive even in real-world scenarios from a survival modelization standpoint while requiring a single communication exchange between server and clients to terminate.

The rest of the paper is organized as follows. Section~\ref{sec:background} provides in-depth background on survival analysis and federated learning. Section~\ref{sec:related_work} reviews the current literature on federated survival analysis, highlighting the applied techniques and healthcare applications. Section~\ref{sec:method} describes the FedSurF++ algorithm. Section~\ref{sec:experiments} analyzes the empirical results obtained, both on simulated federations and on real-world datasets. Finally, Section~\ref{sec:conclusion} summarizes the work.

\section{Background}
\label{sec:background}

In this section, we review the basics of survival analysis and federated learning. First, we define the problem of survival analysis and how it relates to statistical modeling and machine learning (Section~\ref{sec:survival-analysis}). We then categorize and explain the state-of-the-art survival models based on neural networks (Section~\ref{ref:survival_models}) and the most common survival evaluation metrics (Section~\ref{sec:sf_evaluation}). Finally, we introduce federated learning and the techniques for dealing with data heterogeneity in distributed federations (Section~\ref{sec:federated-learning}).

\subsection{Survival Analysis}
\label{sec:survival-analysis}

Survival analysis~\cite{klein2003survival,wang2019machine} is a branch of statistical machine learning concerned with the analysis of time-to-event data, where the event of interest may be death, disease onset, hardware failure, or any other event. The goal of survival analysis is to model the relationship between survival time and predictors related to a particular subject, called features or covariates. The output of survival models is the probability of survival or the risk of experiencing the event over time. Specifically, the survival function 
\begin{equation}
    S(t) = P(T > t)
    \label{eq:s}
\end{equation}
is the probability that an individual will not experience the event, i.e. \emph{survive}, beyond time $t$. It is a non-increasing function, ranging from $1$ at $t=0$ to $0$ for $t\to\infty$. The hazard function 
\begin{equation}
h(t) = \lim_{\delta t \to 0} \frac{P(t \leq T < t + \delta t | T\geq t)}{\delta t}
\end{equation}
is the instantaneous failure rate at time $t$, given that the individual has survived up to $t$. It is a non-negative function that can take any value greater than $0$. The survival function and the hazard function are related as
\begin{equation}
    S(t) = \text{exp}(-H(t))
    \label{eq:s_to_h}
\end{equation}
where $H(t) = \int_{0}^{t} h(\tau) \,d\tau$ is the cumulative hazard function. Each of these functions can be estimated using statistical and machine learning techniques starting from a survival dataset. A survival dataset is a set of triplets 
\begin{equation}
    D=\{(\mathbf{x}_i,\delta_i,t_i)\}_{i=1}^N
\end{equation}
such that 
\begin{itemize}
    \item $\mathbf{x}_i \in \mathbb{R}^d$ is a $d$-dimensional real-valued feature vector.
    \item $\delta_i \in \{0, 1\}$ is an event indicator set to $1$ if the $i$-th subject experienced the event and set to $0$ if the sample is censored instead.
    \item $t_i > 0$ is the event time if $\delta_i = 1$ or the censoring time if $\delta_i = 0$.
\end{itemize}

\subsection{Survival Models}
\label{ref:survival_models}

There are different types of survival models, depending on the assumptions made about the form of the survival or hazard function. 
Non-parametric models make no assumptions about the shape of the survival or hazard function and rely on empirical data aggregations without considering feature vectors. 
These models are easy to calculate and provide unbiased estimates of the survival function. 
These models are most useful for data exploration and visualization purposes. 
Examples of non-parametric models include Kaplan-Meier (KM)~\cite{kaplan1958nonparametric} and Nelson-Aalen~\cite{nelson1972theory,aalen1978nonparametric}. 
Specifically, the KM estimator calculates the cumulative probability of survival based on data by successively multiplying the probabilities of survival at each unique event time.
In particular, for each unique event time $t_j$ in the set $T_D = \{t_j: (\mathbf{x}_i,\delta_i,t_j) \in D\}$, KM counts the number of observed events $d_j$ and the number of samples $r_j$ that are still at risk. 
Then, the KM estimator calculates the survival function $\hat{S}(t)$ by cumulatively multiplying these survival probabilities for all time points preceding $t$, as
\begin{equation}
\hat{S}(t) = \prod\limits_{j: t_j < t} \left(1 - \frac{d_j}{r_j} \right).
\end{equation}
This way, the KM estimator encapsulates the intuition that the probability of surviving up to a given time is the cumulative product of the probabilities of surviving each preceding moment.

Semi-parametric models decompose hazard functions into a common baseline hazard $h_0(t)$ and a subject-related risk function $\phi(t, \mathbf{x})$. The resulting hazard function is calculated as
\begin{equation}
    h(t | \mathbf{x}) = h_0(t) \cdot \phi(t,\mathbf{x}).
\end{equation}
One of the most widely used semi-parametric models is the Cox proportional hazard model~\cite{cox1972regression}. In this model, the risk function for an individual with a feature vector $\mathbf{x}$ is given by
\begin{equation}
    \phi(\mathbf{x}) = \text{exp}\left(\sum\limits_{j=0}^d \beta_j \cdot x_j\right)
\end{equation}
where $\beta$ is the vector of regression coefficients measuring the effect of each element $x_j$ of $\mathbf{x}$ on the hazard function. Cox models are based on the proportional hazard assumption. This assumption states that the hazard ratio between different subjects is constant over time, i.e., the effect of $\mathbf{x}$ on the hazard does not depend on $t$. Cox models are trained using the partial log-likelihood loss function, given by
\begin{equation}
    L(\beta) = \sum\limits_{i=1}^N \delta_j\left[\beta^T\mathbf{x}_i - \text{log}\left(\sum\limits_{j \in R_i} \text{exp}\left(\beta^T\mathbf{x}_j\right)\right)\right]
\end{equation}
where $R_i$ is the set of subjects at risk at time $t_i$. For this reason, this function is \emph{non-separable}, i.e., it requires access to all available samples in the dataset to be evaluated. Cox models have several advantages. First, they are easy to interpret and require little computation due to their linear nature. These models can also incorporate feature vectors without explicitly modeling the baseline hazard. However, they rely on the assumption of proportional hazards, which may not hold for large datasets. The lack of an explicit survival function may also be a disadvantage in studies where risk ratios are not relevant.

Non-linear models use flexible functions, such as neural networks or splines, to capture complex and non-linear relationships between survival times and feature variables. These models have the highest modeling power because they can approximate any survival or hazard function. However, they require large amounts of data and computational resources for training and optimization. They also suffer from poor interpretability and a high risk of overfitting, especially when data pools are small. 

Among the nonlinear models, DeepSurv~\cite{katzman2018deepsurv} is an extension of the Cox proportional hazards model.
DeepSurv replaces the linear risk function, common in traditional survival models, with a single-output neural network. 
This allows for a more flexible representation of complex interactions between covariates, handling nonlinear relationships in high-dimensional data that are challenging for traditional models. 
DeepSurv relies on the same partial log-likelihood loss function as the Cox model, but the risk scores are predicted by a deep neural network rather than a linear function.

DeepHit~\cite{lee2018deephit}, on the other hand, is a neural-based architecture that focuses on the analysis of multiple concurrent events in survival analysis. 
It does this through time discretization, where each interval is modeled as a multi-class classification problem using sigmoid activations.
Each class is tailored to identify the occurrence of one of the concurrent events.
This enables DeepHit to capture complex patterns in time-to-event data, particularly when there are multiple concurrent event types to consider.

Neural Multi-Task Logistic Regression (N-MTLR)~\cite{fotso2018deep} is a non-linear extension of the Multi-Task Logistic Regression (MTLR) model~\cite{yu2011learning}. 
The MTLR model is a discrete-time survival model that calculates the likelihood of an event happening in each time interval using a separate logistic regression model for each time interval.
Similarly to DeepHit, the classification task is tailored to identify whether an event occurred in a specific time interval.
N-MTLR extends MTLR by incorporating a nonlinear predictor for each time bin, based on neural networks, mapping input covariates to a single output. 
These binned outputs are then combined using a softmax function to obtain survival estimates for each time bin.

Nnet-Survival~\cite{gensheimer2019scalable,kvamme2021continuous}, also known as Logistic Hazard, is a model that leverages the discrete formulation of survival problems to model discrete hazard functions. 
Again, the method involves breaking down the survival problem into a series of binary classification tasks, each representing the risk of event occurrence in a particular time interval. 
This allows for flexibility in modeling time-varying effects and interactions.

Piecewise-Constant Hazard (PC-Hazard)~\cite{kvamme2021continuous,bender2021general}, as the name suggests, estimates the hazard function as a piecewise constant. 
This model assumes that the hazard, or the risk of an event happening, remains constant within certain time intervals, but can change between intervals. 
This assumption simplifies the model to a series of regression problems, allowing it to take advantage of existing machine learning techniques. 
The neural component of the model maps covariates to a finite number of outputs, corresponding to the hazard in each interval.
In this way, while the output of the neural network has a discrete size, the resulting survival function is still continuous and can be computed using Equation~\eqref{eq:s_to_h}, resulting in a series of piecewise exponential functions.

Finally, Random Survival Forests (RSFs)~\cite{ishwaran2008random} are a class of ensemble-based models that use survival trees to estimate the cumulative hazard function $H(t)$. They follow the same principle as random forests for classification and regression~\cite{breiman2017classification}, where a large number of binary trees are grown using bootstrap samples of the data. The main difference lies in the node-splitting technique, which maximizes the hazard difference between the child nodes. The resulting hazard function is obtained by averaging the hazard functions of the terminal nodes across all trees, which are computed using the Nelson-Aalen estimators~\cite{nelson1972theory,aalen1978nonparametric} on the leaf samples.

\subsection{Survival Metrics}
\label{sec:sf_evaluation}

The Concordance Index (C-Index), the Integrated Brier Score (IBS), and the Cumulative Area-Under-the-Curve (Cumulative AUC) are the most used metrics to evaluate survival models. The C-Index~\cite{uno2011c} is a measure of the agreement between the predicted and true survival outcomes for a pair of samples. The predicted outcome is the estimated survival probability or risk score for a given time point, and the true outcome is the actual survival time or event status (1 if the event has occurred and 0 otherwise). A pair of samples is comparable if at least one of them has experienced the event of interest. The C-Index is calculated as the ratio of concordant pairs to comparable pairs. A pair is concordant if the sample with the higher predicted outcome survives longer than the sample with the lower predicted outcome. The C-Index ranges from 0 to 1, where 0.5 is a random prediction and 1 is a perfect prediction. The C-index reflects the discriminative power of the model, i.e., its ability to rank samples according to their actual survival times.

The Brier Score~\cite{graf1999assessment} (BS) is a measure of the accuracy of the predicted survival probability for a sample at a given time. The Brier Score is calculated as the squared difference between the true survival status (1 if the event has occurred and 0 otherwise) and the predicted survival probability of the model at that time. The Brier Score ranges from 0 to 1, where 0 indicates a perfect prediction and 1 indicates a completely incorrect prediction. A random guessing model would have a BS of 0.25. Thus, the lower the BS, the better the model. The Brier score reflects also the calibration of the model, i.e., its ability to estimate the correct survival probabilities for each sample. The Integrated Brier Score (IBS) is a measure of the overall calibration of the model over time. The IBS is calculated as the average of the Brier scores over a series of time points. The IBS also ranges from 0 to 1, where 0 indicates a perfect prediction and 1 indicates a completely wrong prediction. 

In survival analysis, the evaluation of the AUC for classification can be extended to time-varying outcomes~\cite{sksurv}. In particular, the time-dependent AUC defines a time interval and compares the predicted survival probability at the beginning of the interval with the observed event status within the interval. Samples that are censored within or before the interval are considered negative cases. The Cumulative AUC is a summary measure that integrates the time-dependent AUC over time. The Cumulative AUC ranges from 0 to 1, with 1 indicating perfect prediction.

Survival metrics can account for the censoring distribution by applying the Inverse Probability of Censoring Weighting (IPCW)~\cite{uno2011c,robins1992recovery}. The IPCW assigns a weight to each sample based on the inverse probability of being censored at a given time point. The weight reflects how representative the sample is of the underlying population at that time point. Samples with a higher probability of being censored are assigned higher weights, and vice versa. IPCW weighting can help to reduce the bias introduced by censored samples in the resulting metrics.

\subsection{Federated Learning}
\label{sec:federated-learning}

Federated Learning (FL)~\cite{li2020federated,kairouz2021advances} is a distributed machine learning paradigm that allows multiple clients to collaboratively train a model without sharing their private data. In FL, data remain on the devices where they are generated, and only model updates are communicated to a central server that coordinates the learning process. This approach contrasts with traditional centralized machine learning techniques, where all local data are uploaded to a single server, as well as more classical decentralized approaches, which often assume that local data are identically distributed. Federated learning allows multiple actors to build a shared machine learning model without sharing data while addressing security and data access rights. In addition, a model trained on distributed heterogeneous data is representative of a large portion of the population. By processing data mostly at the edge, federated learning can reduce latency, power consumption, and communication costs compared to explicitly sharing data with a central server.

In a typical FL setting, there are $K$ clients, each holding a local dataset $D_k$. The goal of a FL algorithm is to learn a set of model parameters $w$ that minimize a global loss function $L$. This function is the weighted sum of local loss functions $L_k$ computed by each client $k$ on their own data $D_k$. Each contribution is weighted in proportion to the number of samples stored in each database $D_k$. Federated Averaging (FedAvg)~\cite{mcmahan2017communication} is the first algorithm proposed to optimize $L$. FedAvg works in rounds, where each round consists of three steps: a broadcast step, a training step, and an aggregation step. In the broadcast step, the server selects a subset of clients and sends them the current model parameters $w$. In the training step, each of these clients trains the model with parameters $w$ on its local data for a small number of epochs, obtaining a new set of parameters $w'_k$. These parameters are then sent back to the server. Finally, in the aggregation step, the server updates the global parameters by taking a weighted average of the updates $w'_k$ received from the clients and repeats the process until convergence.

While FedAvg achieves good performance in simulated settings, it faces several challenges when applied to real-world scenarios where heterogeneity is prevalent both in terms of computational resources and data distribution across clients~\cite{li2020federated}. For example, some clients may have slower computation time or connectivity, resulting in missed updates during federated training. For this purpose, several asynchronous frameworks have been developed that are reliable for stragglers~\cite{chen2020asynchronous}. In addition, data distributions among clients may not be independent and identically distributed (IID), leading to biased or inaccurate updates that can affect the global model quality. To address this issue, some works propose regularization techniques that reduce the discrepancy between local and global models~\cite{reddi2020adaptive,wang2021field,li2020fedprox,karimireddy2020scaffold,acar2021federated}. 

As data heterogeneity is one of the key challenges in federated networks, test and simulation environments are essential for federated learning research, as they allow to evaluate and compare different algorithms under different realistic settings. Several benchmarking methods have been developed for this purpose. LEAF~\cite{caldas2018leaf} provides a collection of heterogeneous datasets for standard machine learning tasks such as image classification and next character prediction. SGDE~\cite{lomurno2022sgde} generates synthetic datasets from privacy-preserving data generators that learn the characteristics of the client's data. Other works~\cite{hsu2019measuring,li2022federated} investigate data splitting techniques, based on the Dirichlet distribution, that adjust the degree of heterogeneity in federated classification datasets.

\section{Related Work}
\label{sec:related_work}

Machine learning methods have significantly advanced healthcare applications, revolutionizing disease diagnosis, prognosis, and treatment~\cite{litjens2017survey,frizzell2017prediction,yue2018deep,piccialli2021survey}. Nevertheless, the application of these methods often requires the use of extensive datasets, which necessitates a careful balance between data utilization and patient privacy preservation.
To answer this issue, Federated Learning (FL) has emerged as a promising approach for large-scale healthcare applications~\cite{rieke2020future}. Leveraging distributed data while maintaining data privacy, FL models have outperformed traditional statistical methods in predicting outcomes based on medical data~\cite{xu2021federated,sheller2020federated,brisimi2018federated}. 

In this context, federated survival analysis models the time to event of interest (such as death, disease, or failure) in a population, where data is distributed across multiple institutions. This field bridges privacy-preserving federated training with conventional survival analysis techniques. Federated survival analysis has proved to be effective, particularly in oncology, contributing to more robust and privacy-preserving predictive models~\cite{andreux2020federated,truly2021froelicher,chowdhury2021review,larynx2022hansen}. Beyond cancer research, other studies have also applied federated survival analysis to examine stroke events~\cite{learning2020duan} and COVID-19 survival rates~\cite{wang2022survmaximin}.

Much effort has been devoted to federated Cox models~\cite{lu2015webdisco,andreux2020federated,learning2020duan,dssurvival2022banerjee,verticox2022dai,larynx2022hansen,accurate2022kamphorst,federated2022masciocchi,wang2022survmaximin,imakura2023dccox,zhang2023federated}. In fact, the Cox proportional hazard model is one of the most prominent models from classical survival analysis that is easy to interpret, fast to compute, and does not require explicit modeling of the baseline hazard function. However, as explained in Section~\ref{sec:background}, the partial log-likelihood is not separable. This is a serious problem in federations where data are confidential, as clients are unable to effectively compute hazard rates for the entire data. Many alternative formulations to the standard Cox model have been proposed. For example, the authors of~\cite{andreux2020federated} propose a discrete extension of the proportional Cox model to formulate survival analysis as a classification problem with a separable loss function. The method in~\cite{zhang2023federated} is also based on discretization but takes into account the effects of time-varying covariates. In~\cite{learning2020duan}, patient-level data from one client is combined with aggregated information from the other clients to construct a surrogate likelihood function that approximates the Cox partial likelihood function obtained using all available patient-level data. Cox models have been adapted for vertically partitioned data, where data samples from the same patients are stored in different institutions~\cite{verticox2022dai,accurate2022kamphorst,imakura2023dccox}. In particular, VERTICOX~\cite{verticox2022dai} is an algorithm based on the ADMM framework~\cite{boyd2011distributed} that obtains global model parameters in a distributed manner by computing and exchanging intermediate statistics, achieving an accuracy similar to that of a centralized Cox model.

Federated implementations of classical survival models are not limited to Cox. In~\cite{truly2021froelicher}, the authors propose FAMHE, a federated system that allows privacy-preserving estimation of Kaplan-Meier models. Regarding nonlinear models, the authors of~\cite{rahimian2022practical} propose a method to improve the performance of nonlinear federated survival models with differential privacy by adding a post-processing step that adjusts the magnitude of the average noisy parameter update and facilitates model convergence. In~\cite{federated2022lu}, weakly supervised attention modules are used to estimate discrete survival rates. FedPseudo~\cite{rahman2022fedpseudo} uses pseudo values as surrogate labels for federated deep learning models. Finally, FedSurF~\cite{archetti2023federated} leverages federated ensemble learning to construct random survival forests from distributed survival data.

Privacy is a crucial aspect of survival applications, as patient-level data are often sensitive and confidential. Several works have used differential privacy~\cite{dwork2008differential} to protect survival models against inference attacks~\cite{federated2022lu,rahimian2022practical}. Alternatively, some works have relied on secure multi-party computation (SMPC), which allows computation on distributed data without revealing individuals' information. For example, SMPC has been applied to the Newton-Raphson algorithm to optimize the partial log-likelihood of distributed Cox models by computing intermediate statistics~\cite{accurate2022kamphorst}. Another SMPC protocol, SecureFedYJ~\cite{securefedyj2022marchand}, allows the Yeo-Johnson transformation to be applied to vertically-partitioned data while preserving privacy. FAHME~\cite{truly2021froelicher} uses multiparty homomorphic encryption to estimate distributed Kaplan-Meier models. Finally, in~\cite{imakura2023dccox}, a Cox model is designed for horizontal and vertical federated learning exploiting a privacy-preserving subspace projection technique that allows each local institution to obtain a secure approximation of the model parameters, survival curves, and statistics such as p-values.

Federated learning has been applied to genomic analysis for cancer survival and recurrence studies~\cite{andreux2020federated,federated2022lu,terrail2022flamby}, using data from The Cancer Genome Atlas (TCGA) project. The TCGA Project is a large-scale database from the National Cancer Institute and the National Human Genome Research Institute that molecularly characterizes 33 types of cancer by collecting genomic, epigenomic, transcriptomic, and proteomic data that are publicly available to researchers. Among the works based on the TCGA project, FLamby (Federated Learning AMple Benchmark of Your cross-silo strategies)~\cite{terrail2022flamby} is a collection of cross-silo federated learning datasets for healthcare applications. One of the datasets in the Flamby suite, Fed-TCGA-BRCA, consists of genomic and clinical data of breast cancer patients from 6 different hospitals. The dataset is naturally partitioned according to the geographic origin of the patients, with each patient assigned to the closest center. 
Finally, to evaluate the performance and compare the results of federated survival analysis methods, \cite{archetti2023heterogeneous} provides two algorithms to split existing survival datasets into heterogeneous federations.

\section{Method}
\label{sec:method}

In this section, we present the proposed extension of the Federated Survival Forest  (FedSurF) algorithm~\cite{archetti2023federated}, called FedSurF++. As the original algorithm, FedSurF++ relies on Random Survival Forests (RSF)~\cite{ishwaran2008random} to build a tree-based ensemble model for survival analysis in a federated learning setting. 
Our approach builds upon prior works in federated ensemble learning~\cite{hauschild2022federated,gencturk2022bofrf}, where the central server merges base models from local ensembles on each client to create a global model. 
Specifically, the FedSurF++ algorithm constructs a RSF on the central server by aggregating the top-performing trees from local RSF models on each client, with an emphasis on the tree sampling strategy. 

\subsection{The FedSurF++ Algorithm}

The FedSurF++ algorithm consists of three steps: \emph{local training}, \emph{tree assignment}, and \emph{tree sampling}. The \emph{local training} and \emph{tree assignment} stages remain unchanged from the original FedSurF algorithm. In particular, during the \emph{local training} step, each client $k$ builds a local RSF $M_k$ from the local data $D_k$. At this point, each local RSF model $M_k$ is a set of $T_k$ survival trees. These binary trees are built with a recursive node-splitting technique inspired by CART~\cite{breiman1984classification} that maximizes the survival difference between samples in child nodes. Tree leaves contain the Nelson-Aalen estimator~\cite{nelson1972theory} of the cumulative hazard resulting from their samples. Each client may tune the RSF hyperparameters to best fit their data distribution and hardware constraints, making local execution feasible and effective. For example, clients with high computational power can train forests with high cardinality, while clients with hardware limitations can lower the number of trees in their local models.

In the \emph{tree assignment} stage, the server determines the number of trees each client is required to send on the server. To this end, the server iteratively increments a client counter $T'_k \leq T_k$ for a number of times equal to the number of desired trees $T$ in the final ensemble $M$. Intuitively, the counter for client $k$ cannot exceed the number of trees $T_k$ in their local model $M_k$. At each iteration, a client counter $T'_k$ is incremented with a probability proportional to $N_k=|D_k|$. This is to promote the selection of trees coming from clients that have larger datasets. This way, FedSurF++ promotes trees trained on larger data samples, which are likely to be more representative of the entire population. This procedure is inspired by the weighted updates of FedAvg~\cite{mcmahan2017communication} that assign a weight proportional to the local dataset cardinality when aggregating model parameters coming from different clients.

Finally, in the \emph{tree sampling} stage, each client samples $T'_k$ trees to be shared with the server. We introduce three new sampling strategies that are proportional to the Concordance Index (C-Index), the Concordance Index with IPCW weighting (C-Index-IPCW), and the Cumulative Area-Under-the-Curve (Cumulative AUC). This is an extension with respect to the original FedSurF, which is limited to sample trees according to the inverse of the Integrated Brier Score (IBS). Each of these metrics is discussed in Section~\ref{sec:sf_evaluation}. Given one of these metrics, clients evaluate each local tree, obtaining a set of estimations $\{\text{Metric}_j\}_{j=1}^{T_k}$. At this point, each client selects $T'_k$ trees with a probability proportional to the chosen metric. In order to differentiate between sampling strategies, we adopt different names for our algorithm. Specifically, if clients choose to use a uniform sampling strategy, the method is referred to as FedSurF. For each of the metric-based sampling strategies, instead, we denote the method as FedSurF-Metric, where Metric represents the chosen performance measure. Section~\ref{sec:experiments} collects experiments comparing the uniform sampling strategy (FedSurF) and the strategies proportional to the C-Index (FedSurF-C), the C-Index-IPCW (FedSurF-C-IPCW), the inverse IBS (FedSurF-IBS), and the Cumulative AUC (FedSurF-AUC). 

While FedSurF++ is a relatively straightforward extension of the original FedSurF algorithm, it allows us to delve deeper into how tree-sampling methods affect the corresponding metric in the final model. Our experimental findings suggest that using the least expensive evaluation metric can still produce a high-performing model. Consequently, trees can be sampled using the C-Index without IPCW weighting, as in FedSurF-C, to achieve the optimal ensemble model. This has several implications from the algorithmic perspective. First, it allows each client to evaluate concordance locally, without relying on the aggregated statistics of other clients. This way, the number of messages to be shared in a federation is reduced. Second, this simpler metric protects privacy, as cumulated statistics are not shared with the server or any other client in the federation.

In summary, FedSurF++ extends FedSurF with a simple yet natural operation -- allowing the selection of the tree-sampling method -- that results in important implications from the efficiency, communication, and privacy perspectives.  
The pseudocode of FedSurF++ is presented in Algorithm~\ref{alg:fedsurf}. 

\begin{algorithm}[t]
\caption{FedSurF++ Algorithm}
\label{alg:fedsurf}
\begin{algorithmic}
\Function{FedSurF-Client}{$D_k$}
\State \(\triangleright\) \emph{Local training}
\State Tune parameters of local RSF $M_k$ using cross-validation.
\State Train local RSF $M_k$ on $D_k$.
\State Send the number of local trees $T_k$ to the central server.
\State \(\triangleright\) \emph{Tree sampling}
\For{$j = 1$ to $T_k$}
    \State Compute $\text{Metric}_j$ for tree $j \in M_k$.
\EndFor
\State Receive $T'_k$, the number of trees to send back to the server.
\State Select $T'_k$ trees using probabilities proportional to $\text{Metric}_j$.
\State Send selected trees to the server.
\EndFunction
\State
\Function{FedSurF-Server}{$T$}
\State \(\triangleright\) \emph{Tree assignment}
\State Receive $T_k$ from each client $k$.
\State Compute $T'_k$ for each client $k$ according to $T_k$ and $T$.
\State Send $T'_k$ to each client $k$
\State \(\triangleright\) \emph{Model construction}
\State Receive $T'_k$ trees from each client $k$.
\State Construct the final model $M$ by aggregating $T$ trees.
\State \textbf{Return:} Random survival forest $M$.
\EndFunction
\end{algorithmic}
\end{algorithm}

\subsection{Computational Complexity}

Deriving an accurate estimate of the computational complexity related to federated algorithms is a complex task, as many factors that arise in real-world scenarios are difficult to integrate into the analysis.
However, we can derive a rough estimate of the computational complexity of FedSurF++ to assess its strengths and limitations from a scalability perspective.

The training time complexity of RSFs primarily involves the number of trees in the forest $T$, the number of samples $N=|D|$, the number of features $F$, and the depth of the trees (which can be, at most, $\log(N)$ in a balanced tree scenario). 
RSFs use log-rank tests~\cite{bland2004logrank} to determine the best split at each node. Log-rank tests compare the survival distributions of two groups to determine if they are statistically different, which is an $O(N)$ operation. As this operation is performed at each node, it multiplies by the number of candidate features for splitting, $\sqrt{F}$. If the tree is fully expanded, the number of internal nodes in a binary tree is, at most, $N - 1$. Therefore, the overall complexity related to node splitting is $O\left(T \cdot N^2 \cdot \sqrt{F}\right)$.
At each leaf node, the Nelson-Aalen estimator accounts for a cost of $O(N)$. Since there could be at most $N$ leaves in a fully expanded tree, the overall leaf evaluation complexity is $O\left(T \cdot N^2\right)$.

By combining the complexity of node splitting and leaf computations, the overall training time complexity for RSFs is $O\left(T \cdot N^2 \cdot \sqrt{F}\right)$.
However, in practice, trees are not usually fully grown, as they are pruned or have a maximum depth, and samples are bootstrapped in each tree. The complexity of FedSurF++ is comparable to a single RSF execution, as it requires a single communication round, once the forests are trained in parallel on the clients. To complete the analysis, Section~\ref{sec:sf_training} collects empirical time executions for RSFs and neural-based models.

\section{Experiments}
\label{sec:experiments}

This section collects the experiments based on which we compare the performance of FedSurF++ with other neural-based models from the state of the art in federated survival analysis. We present two sets of experiments. The former focuses on simulated federations, and the latter focuses on real-world federations. In particular, Section~\ref{sec:simulated_federations} reports the experiments on federations with splits simulated by the label-skewed splitting algorithm~\cite{archetti2023heterogeneous}. Instead, Section~\ref{sec:real_world_federations} collects the experiments on Lombardy Heart Failure~\cite{Mazzali2016MethodologicalIO} and Fed-TCGA-BRCA~\cite{terrail2022flamby}, which are based on real-world data splits.

\subsection{Experiments on Simulated Federations}
\label{sec:simulated_federations}

This section covers the experiments related to simulated uniform and heterogeneous data splits of existing survival datasets.

\subsubsection{Datasets}
\label{sec:sf_datasets}

The following survival datasets are commonly used to evaluate non-federated survival methods. From these, we conduct experiments on simulated federations. Table~\ref{tab:survival-datasets} summarizes the statistics of these datasets. \begin{table}[t]
    \centering
    \caption{Survival datasets for simulated federation studies.}
    \label{tab:survival-datasets}
    
    \begin{tabularx}{\linewidth}{@{} l Y Y Y Y @{}}
        \toprule
        \textbf{Dataset} & \textbf{Samples} & \textbf{Censored} & \textbf{Covariates} \\
        \midrule
        WHAS500~\cite{hosmer_applied_2008} & 461 & 38\% & 16\\
        GBSG2~\cite{schumacher1994randomized} & 686 & 44\% & 8\\
        METABRIC~\cite{pereira_somatic_2016,katzman2018deepsurv} & 1904 & 58\% & 8\\
        NWTCO~\cite{breslow_design_1999} & 4028 & 14\% & 8\\
        FLCHAIN~\cite{therneau2023survival} & 7874 & 28\% & 10\\
        \bottomrule
    \end{tabularx}
\end{table}

\begin{itemize}
    \item The Worcester Heart Attack Study (WHAS500) dataset~\cite{hosmer_applied_2008} contains data on 461 patients who experienced acute myocardial infarction. The data were collected during the first hospitalization and included 16 covariates. The outcome of interest is survival time after the event.
    \item The German Breast Cancer Study Group (GBSG2) dataset \cite{schumacher1994randomized} examines the effects of hormone treatment for breast cancer in 686 women. The outcome of interest is time to cancer recurrence. The dataset includes 8 covariates, such as age, menopausal status, tumor grade and size, and hormone levels.
    \item The Molecular Taxonomy of Breast Cancer International Consortium (METABRIC) dataset~\cite{pereira_somatic_2016,katzman2018deepsurv} is a Canada-UK project that provides survival data for 1904 patients with breast cancer. The dataset comprises clinical attributes, gene expression profiles, copy number variations, and single nucleotide polymorphisms as covariates.
    \item The National Wilm's Tumor Study (NWTCO) dataset~\cite{breslow_design_1999} consists of 4028 observations on 8 covariates for patients with Wilm's tumor, a rare type of kidney cancer that primarily affects children. The covariates include histology status, disease stage, and other factors. The outcome of interest is time to relapse.
    \item The FLCHAIN survival dataset~\cite{dispenzieri_use_2012} contains subjects from a study concerning mortality rates of serum-free light chain (FLC). The original data come from the residents of Olmsted County, Minnesota, with more than 50 years. The dataset has 7874 samples and 10 features.
\end{itemize}

\subsubsection{Simulated Federations}

The datasets from Section~\ref{sec:sf_datasets} are used to simulate federated datasets by assigning each sample to a particular client in the federation. First, each dataset is split into a training set and a test set by randomly selecting 70\% of the total samples for training and the remaining 30\% for testing.

\begin{figure*}[t!]
    \centering
    \includegraphics[width=\textwidth]{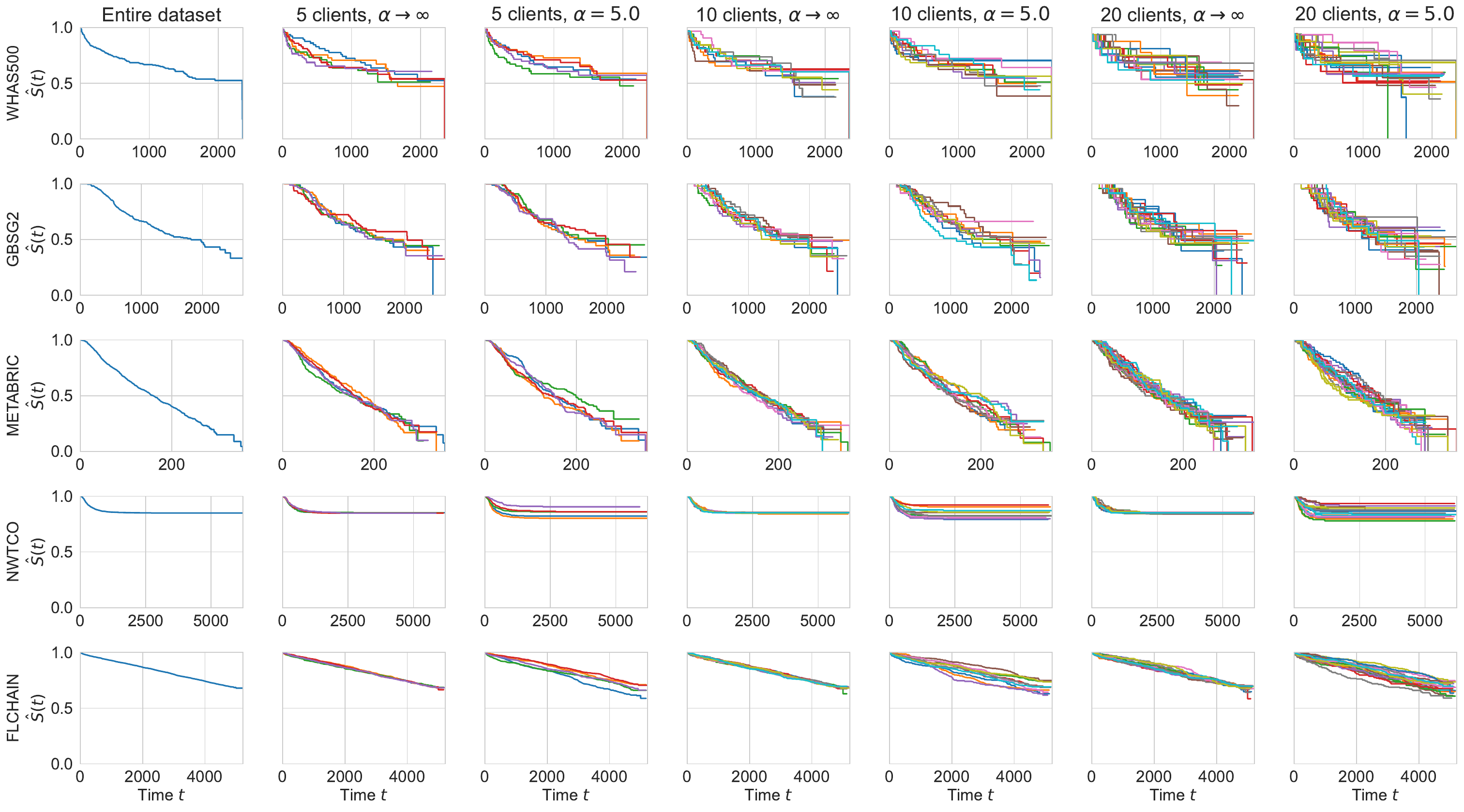}
    \caption{Kaplan-Meier estimators $\hat{S}(t)$ for datasets of simulated federations. The first row shows KM estimators for the entire dataset, while the second, third, and fourth rows depict KM curves for 5, 10, and 20 clients, respectively.}
    \label{fig:sim_km}
\end{figure*}

Then, each sample in the training set is assigned to one of the clients in the federation. We assume that there are 5, 10, or 20 cooperating institutions in each simulated experiment, i.e., $K=5, 10$, or $20$ depending on the considered setting. Sample assignment is performed by the label-skewed splitting algorithm~\cite{archetti2023heterogeneous}. This algorithm is based on the Dirichlet distribution as in~\cite{hsu2019measuring,li2022federated}. The aim is to create unbalanced distributions of times and events across clients. In fact, it is important to test federated algorithms in heterogeneous contexts, as non-identical data distributions can affect the performance and convergence speed of federated algorithms. The label-skewed splitting algorithm has a hyperparameter $\alpha$ that controls the degree of heterogeneity among clients. Smaller values of $\alpha$ result in more heterogeneous distributions. We set $\alpha\rightarrow\infty$ to simulate a federation with uniformly split data and $\alpha=5$ to simulate a federation with heterogeneous data distribution. The label distributions for each client are shown in Figure~\ref{fig:sim_km}. We plot the Kaplan-Meier estimator for each client in the federation. By inspecting the estimators, federations with $\alpha\rightarrow\infty$ exhibit similar label distributions across clients, while for $\alpha=5$ distributions differ.

In addition, 30\% of the local samples on each client are reserved for validation and hyperparameter tuning. During our simulations, clients are assumed to be always available and communication packets are never lost.

\subsubsection{Baseline Models}
\label{sec:sf_models}

FedSurF++ is compared with the six state-of-the-art survival models described in Section~\ref{ref:survival_models}. The first is the Cox proportional hazards model (CoxPH)~\cite{cox1972regression}, which uses the Nelson-Aalen estimator~\cite{nelson1972theory} to estimate the baseline hazard. A nonlinear extension of the Cox model, DeepSurv~\cite{katzman2018deepsurv}, is also included. The other models are discretized survival models based on neural networks: DeepHit~\cite{lee2018deephit}, Neural Multi-Task Logistic Regression (N-MTLR)~\cite{fotso2018deep}, and Nnet-Survival~\cite{gensheimer2019scalable}. Finally, we consider Piecewise-Constant Hazard (PC-Hazard)~\cite{kvamme2021continuous}, which is a non-proportional hazard neural-based model that provides a continuous estimation of the survival function.

The neural network architectures of DeepSurv, DeepHit, N-MTLR, Nnet-Survival, and PC-Hazard consist of two fully connected layers of 32 neurons each with ReLU activation functions. We also add a dropout layer with a probability of 10\% to prevent overfitting. The output of DeepSurv is a scalar obtained by a linear transformation while the other models have 10 outputs corresponding to different discretization instants. Specifically, DeepHit, N-MTLR, and Nnet-Survival produce 10 survival probabilities, while PC-Hazard produces 10 discrete hazard values that are converted to survival probabilities using Equation~\ref{eq:s_to_h}.

\subsubsection{Training}
\label{sec:sf_training}

Each model is evaluated in three settings: \emph{Global}, \emph{Local}, and \emph{Federated}. In the \emph{Global} setting, it is assumed that the entire survival dataset is centralized in a single node, and a single model is trained. This setting does not require federated learning and serves as an empirical upper bound to assess the performance loss due to data distribution.

The \emph{Local} setting involves clients training their models only on their local data without participating in the federation. The average performance of the local models is reported as an empirical lower bound, which is targeted for improvement using federated learning. It is expected that joining a federated learning algorithm would benefit the clients in terms of model performance.

In the \emph{Federated} setting, multiple clients collaborate in a federated learning procedure.
To achieve the most effective baseline model training, we performed a comparative analysis between the widely used Federated Averaging algorithm (FedAvg)~\cite{mcmahan2017communication}, and FedProx~\cite{li2020fedprox}, an alternative algorithm aimed at enhancing generalization in heterogeneous federations.
Our study employed the five datasets listed in Table~\ref{tab:survival-datasets}, examining three distinct client configurations ($K=5,10,20$). We ran both FedAvg and FedProx training on each of the six baseline models (CoxPH, DeepSurv, DeepHit, N-MTLR, Nnet-Survival, and PC-Hazard), culminating in a total of 90 direct FedAvg versus FedProx comparisons based on the dataset, model, and client number.
Each of these 90 pairings was repeated five times, followed by a t-test analysis. Notably, only a fraction (5.6\%) displayed a statistically significant performance variation in concordance index between FedAvg and FedProx. Consequently, we chose to utilize the standard FedAvg algorithm for training each neural model in the subsequent experiments.
Furthermore, we assumed that each proportional hazard model, i.e., Cox and DeepSurv, has access to a global Kaplan-Meier estimate of the survival data. 

Federated averaging is implemented using the Flower library~\cite{beutel2020flower} for Python and run for 150 rounds, allowing each client to execute 2 local epochs for each round. 
The best model parameters are selected based on the highest concordance index on the validation set of each client. 
RSFs are implemented with scikit-survival~\cite{sksurv}, and neural-based models are implemented with PyCox~\cite{kvamme2019time}. 
The Adam optimizer with a learning rate of 0.01 is employed to train the neural-based models.

\begin{figure}[t!]
    \centering
    \includegraphics[width=\linewidth]{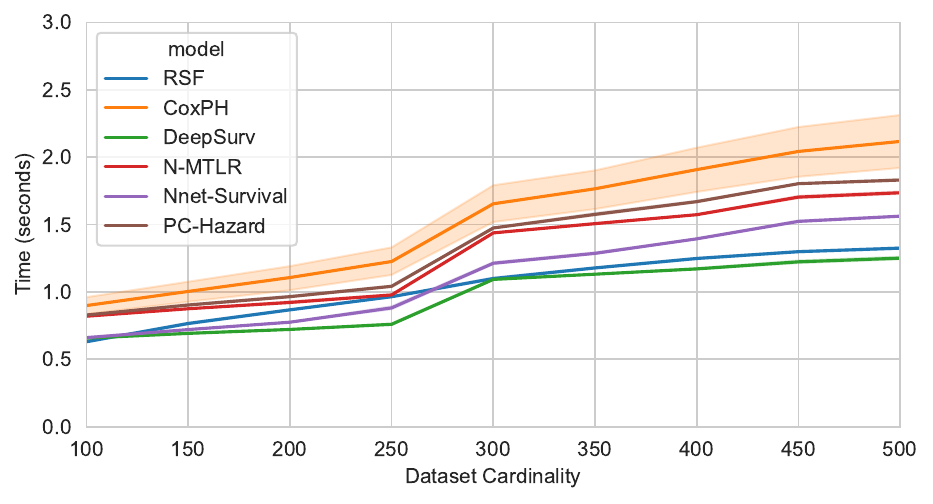}
    \caption{Execution time for each model on several cuts of the GBSG2 dataset. For neural models, time refers to 300 epochs. Results are averaged over 100 runs.}
    \label{fig:time}
\end{figure}

Figure~\ref{fig:time} collects the average execution time of each model for centralized training. Results show that RSFs have a comparable execution time to neural-base models.

\subsubsection{RSF Parameters}
\label{sec:rsf_params}

We optimized the RSF parameter configuration for each dataset adopting a cross-validation approach. 
The most impactful parameters we discovered were the number of estimators $T$ and the maximum tree depth $d$.
Upon conducting a grid search, we determined the optimal $T$ values within the range of 100 to 4000 for each dataset, analyzed at intervals of 20. 
We found a point of diminishing returns for each dataset beyond which increasing the number of trees did not significantly improve the results. 
Then, we fixed the number of trees beyond this identified threshold.

As for $d$, our investigation considered trees of unrestricted depth and trees with a fixed depth of 1. 
Meanwhile, we retained the default values for other parameters such as the minimum number of samples needed to split an internal node $s$ and the minimum number of samples required for a leaf node $l$. 
Specifically, for the scikit-survival implementation we used, $s$ and $l$ were fixed at 6 and 3, respectively.
The maximum number of features retained for each tree was set to the square root of the total number of features in the dataset. 
Moreover, we did not impose any constraints on the maximum number of leaf nodes per tree.
The parameters determined through this analysis are reported in Table~\ref{tab:params}. These values were then applied across all clients in our experiments.

\begin{table}[t]
    \centering
    \caption{RSF parameters for each dataset.}
    \label{tab:params}
    
    \begin{tabularx}{\linewidth}{@{} l Y Y @{}}
        \toprule
        \textbf{Dataset} & \textbf{$T$} & \textbf{$d$} \\
        \midrule
        WHAS500~\cite{hosmer_applied_2008} & 400 & 1\\
        GBSG2~\cite{schumacher1994randomized} & 700 & 1\\
        METABRIC~\cite{pereira_somatic_2016,katzman2018deepsurv} & 500 & $\infty$\\
        NWTCO~\cite{breslow_design_1999} & 600 & 1\\
        FLCHAIN~\cite{therneau2023survival} & 200 & $\infty$\\
        LombardyHF~\cite{therneau2023survival} & 1000 & $\infty$\\
        Fed-TCGA-BRCA~\cite{therneau2023survival} & 1000 & $\infty$\\
        \bottomrule
    \end{tabularx}
\end{table}

\subsubsection{Evaluation}

The Concordance Index (C-Index-IPCW), the Integrated Brier Score (IBS), and the Cumulative Area-Under-the-Curve (Cumulative AUC) are the metrics used to evaluate our survival models on test splits. We account for the censoring distribution by applying the Inverse Probability of Censoring Weighting (IPCW)~\cite{uno2011c,robins1992recovery}, as described in Section~\ref{sec:sf_evaluation}.

\subsubsection{Simulated Federations Results}
\label{sec:sf_results}

\begin{table*}[ht!]
\centering
\caption{Concordance Index with IPCW weighting (C-Index-IPCW)~\cite{uno2011c,robins1992recovery} for survival models evaluated on simulated federations ($K=10$, $\alpha\rightarrow\infty$). Each C-Index-IPCW is scaled by a factor of 100 for better readability. We report the mean computed over 20 runs. The best results ($\uparrow$) are highlighted in bold. Values marked with * do not exhibit statistically significant differences with FedSurF-C according to Dunn's test with 0.05 significance.}
\label{tab:sim_cid_uni}
    
\begin{tabularx}{\textwidth}{@{} l YYY YYY YYY YYY YYY}
\toprule
& \multicolumn{3}{c}{WHAS500} & \multicolumn{3}{c}{GBSG2} & \multicolumn{3}{c}{METABRIC} & \multicolumn{3}{c}{NWTCO} & \multicolumn{3}{c}{FLCHAIN}\\
\cmidrule(l){2-4}\cmidrule(l){5-7}\cmidrule(l){8-10}\cmidrule(l){11-13}\cmidrule(l){14-16}
Model & \emph{Loc.} & \emph{Fed.} & \emph{Glob.} & \emph{Loc.} & \emph{Fed.} & \emph{Glob.} & \emph{Loc.} & \emph{Fed.} & \emph{Glob.}& \emph{Loc.} & \emph{Fed.} & \emph{Glob.}& \emph{Loc.} & \emph{Fed.} & \emph{Glob.}\\ 
\midrule
\mbox{CoxPH} & 65.0 & 74.5 & 77.8 & 56.9 & 61.2 & 63.7 & 57.8 & 62.0* & 64.6 & 63.5 & 65.9 & 53.4 & 90.7 & 91.7 & 94.2
\\
\mbox{DeepSurv} & 67.8 & 77.4 & 73.3 & 58.3 & \textbf{65.9}* & 65.9 & 60.3 & \textbf{64.5}* & 64.8 & 65.2 & \textbf{69.7}* & 55.0 & 93.3 & \textbf{94.2} & 94.3
\\ 
\mbox{DeepHit} & 66.8 & 76.8 & 75.2 & 56.2 & 63.5* & 64.3 & 57.4 & 62.4* & 62.0 & 52.6 & 69.1* & 71.6 & 92.7 & 93.7* & 94.1
\\ 
\mbox{N-MTLR} & 66.8 & 76.2 & 74.4 & 58.0 & 65.5* & 63.9 & 59.6 & 63.8* & 64.4 & 65.2 & 70.2* & 71.6 & 93.5 & \textbf{94.2} & 94.1
\\ 
\mbox{Nnet-Survival} & 65.0 & 74.6 & 75.9 & 54.9 & 60.8 & 63.5 & 50.3 & 58.5 & 62.7 & 44.3 & 66.7 & 70.2 & 87.8 & 93.9* & 94.2
\\ 
\mbox{PC-Hazard} & 65.1 & 74.7 & 75.5 & 54.9 & 60.6 & 63.6 & 50.4 & 58.9 & 62.9 & 44.8 & 66.5 & 70.2 & 88.1 & 93.8*   & 94.2
\\ 
\midrule
\mbox{FedSurF} & 73.0 & 79.2* & 78.6 & 61.8 & 65.4* & 64.2 & 60.7 & 63.8* & 64.0 & 67.7 & 69.1* & 68.1 & 93.6 & 93.8* & 93.9
\\ 
\mbox{FedSurF-C} & -- & 79.5* & -- & -- & 65.3* & -- & -- & 63.9* & -- & -- & 69.1* & -- & -- & 93.9* & --
\\ 
\mbox{FedSurF-C-IPCW} & -- & 79.4* & -- & -- & 65.6* & -- & -- & 63.8* & -- & -- & 69.1* & -- & -- & 93.8* & --
\\ 
\mbox{FedSurF-IBS} & -- & 79.3* & -- & -- & 65.6* & -- & -- & 63.9* & -- & -- & 69.0* & -- & -- & 93.9* & --
\\ 
\mbox{FedSurF-AUC} & -- & \textbf{79.8}* & -- & -- & 65.4* & -- & -- & 64.0* & -- & -- & 69.1* & -- & -- & 93.8* & --
\\ 
\bottomrule
\end{tabularx}

\end{table*}

\begin{table*}[ht!]
\centering
\caption{Integrated Brier Score (IBS)~\cite{graf1999assessment} for survival models evaluated on simulated federations ($K=10$, $\alpha\rightarrow\infty$). Each IBS is scaled by a factor of 100 for better readability. We report the mean computed over 20 runs. The best results ($\downarrow$) are highlighted in bold. Values marked with * do not exhibit statistically significant differences with FedSurF-C according to Dunn's test with 0.05 significance.}
\label{tab:sim_ibs_uni}
    
\begin{tabularx}{\textwidth}{@{} l YYY YYY YYY YYY YYY}
\toprule
& \multicolumn{3}{c}{WHAS500} & \multicolumn{3}{c}{GBSG2} & \multicolumn{3}{c}{METABRIC} & \multicolumn{3}{c}{NWTCO} & \multicolumn{3}{c}{FLCHAIN}\\
\cmidrule(l){2-4}\cmidrule(l){5-7}\cmidrule(l){8-10}\cmidrule(l){11-13}\cmidrule(l){14-16}
Model & \emph{Loc.} & \emph{Fed.} & \emph{Glob.} & \emph{Loc.} & \emph{Fed.} & \emph{Glob.} & \emph{Loc.} & \emph{Fed.} & \emph{Glob.}& \emph{Loc.} & \emph{Fed.} & \emph{Glob.}& \emph{Loc.} & \emph{Fed.} & \emph{Glob.}\\ 
\midrule
\mbox{CoxPH} & 18.5 & \textbf{16.5}* & 14.8 & 19.1 & 18.0* & 16.8 & 17.3 & 16.4* & 15.8 & 11.3 & 11.1* & 11.8 & 8.4 & 6.7 & 4.1
\\
\mbox{DeepSurv} & 19.4 & \textbf{16.5}* & 21.5 & 19.1 & \textbf{16.6}* & 16.8 & 17.2 & \textbf{15.8}* & 16.2 & 11.3 & \textbf{10.4} & 11.8 & 5.7 & 4.2* & 4.2
\\ 
\mbox{DeepHit} & 19.6 & 17.0* & 16.5 & 20.1 & 18.2* & 17.9 & 17.7 & 17.1* & 16.3 & 14.7 & 10.7* & 10.4 & 6.4 & 4.5* & 4.3
\\ 
\mbox{N-MTLR} & 19.4 & 17.5* & 20.4 & 19.4 & 16.8* & 17.9 & 17.2 & 15.9* & 16.1 & 11.7 & \textbf{10.4} & 10.2 & 4.9 & \textbf{4.1}* & 4.4
\\ 
\mbox{Nnet-Survival} & 22.4 & 18.6 & 17.0 & 21.9 & 18.7 & 17.2 & 22.3 & 18.8 & 16.4 & 16.5 & 10.8* & 10.1 & 7.7 & 4.5* & 4.2
\\ 
\mbox{PC-Hazard} & 22.1 & 18.5 & 17.3 & 21.6 & 19.0 & 17.2 & 22.3 & 22.8 & 16.5 & 16.2 & 10.8* & 10.1 & 7.6 & 4.6 & 4.2
\\ 
\midrule
\mbox{FedSurF} & 18.1 & 17.4* & 17.5 & 18.7 & 18.0* & 18.2 & 17.3 & 16.2* & 16.2 & 11.1 & 11.0* & 11.0 & 4.7 & 4.5* & 4.1
\\ 
\mbox{FedSurF-C} & -- & 17.0* & -- & -- & 17.8* & -- & -- & 16.1* & -- & -- & 11.0* & -- & -- & 4.4* & --
\\ 
\mbox{FedSurF-C-IPCW} & -- & 17.0* & -- & -- & 17.8* & -- & -- & 16.1* & -- & -- & 11.0* & -- & -- & 4.4* & --
\\ 
\mbox{FedSurF-IBS} & -- & 17.0* & -- & -- & 17.9* & -- & -- & 16.1* & -- & -- & 11.0* & -- & -- & 4.3* & --
\\ 
\mbox{FedSurF-AUC} & -- & 17.0* & -- & -- & 17.8* & -- & -- & 16.1* & -- & -- & 11.0* & -- & -- & 4.4* & --
\\ 
\bottomrule
\end{tabularx}

\end{table*}

\begin{table*}[ht!]
\centering
\caption{AUC for survival models evaluated on simulated federations ($K=10$, $\alpha\rightarrow\infty$). Each C-Index-IPCW is scaled by a factor of 100 for better readability. We report the mean computed over 20 runs. The best results ($\uparrow$) are highlighted in bold. Values marked with * do not exhibit statistically significant differences with FedSurF-C according to Dunn's test with 0.05 significance.}
\label{tab:sim_auc_uni}
    
\begin{tabularx}{\textwidth}{@{} l YYY YYY YYY YYY YYY}
\toprule
& \multicolumn{3}{c}{WHAS500} & \multicolumn{3}{c}{GBSG2} & \multicolumn{3}{c}{METABRIC} & \multicolumn{3}{c}{NWTCO} & \multicolumn{3}{c}{FLCHAIN}\\
\cmidrule(l){2-4}\cmidrule(l){5-7}\cmidrule(l){8-10}\cmidrule(l){11-13}\cmidrule(l){14-16}
Model & \emph{Loc.} & \emph{Fed.} & \emph{Glob.} & \emph{Loc.} & \emph{Fed.} & \emph{Glob.} & \emph{Loc.} & \emph{Fed.} & \emph{Glob.}& \emph{Loc.} & \emph{Fed.} & \emph{Glob.}& \emph{Loc.} & \emph{Fed.} & \emph{Glob.}\\ 
\midrule
\mbox{CoxPH} & 66.4 & 75.0 & 79.3 & 61.8 & 69.5 & 74.4 & 60.1 & 65.8 & 69.0 & 65.2 & 67.2* & 54.0 & 92.3 & 93.2 & 95.6
\\
\mbox{DeepSurv} & 68.0 & 77.4* & 73.0 & 63.0 & 74.8* & 73.7 & 63.2 & 69.0* & 68.9 & 66.1 & 70.9* & 55.1 & 94.7 & 95.6* & 95.9
\\ 
\mbox{DeepHit} & 65.7 & 76.0 & 74.5 & 58.9 & 71.5 & 70.4 & 59.8 & 69.0* & 69.8 & 52.7 & 73.0 & 74.6 & 93.9 & 95.5* & 95.4
\\ 
\mbox{N-MTLR} & 67.4 & 75.9 & 73.4 & 62.3 & 73.4* & 72.1 & 63.3 & \textbf{70.7}* & 72.3 & 67.6 & \textbf{71.9} & 74.6 & 94.7 & 95.8* & 95.7
\\ 
\mbox{Nnet-Survival} & 62.2 & 72.6 & 75.7 & 57.4 & 68.2 & 72.7 & 48.9 & 61.7 & 69.5 & 36.7 & 68.5* & 72.1 & 88.9 & 95.4* & 95.9
\\ 
\mbox{PC-Hazard} & 62.3 & 72.1 & 74.5 & 57.7 & 67.7 & 73.0 & 48.8 & 60.5 & 69.5 & 37.2 & 67.4* & 71.8 & 88.9 & 94.9 & 95.3
\\ 
\midrule
\mbox{FedSurF} & 73.6 & 79.7* & 80.0 & 68.4 & 74.9* & 73.7 & 64.8 & 69.9* & 71.0 & 68.2 & 69.8* & 68.5 & 94.9 & 95.6* & 96.1
\\ 
\mbox{FedSurF-C} & -- & 79.9* & -- & -- & 74.9* & -- & -- & 70.1* & -- & -- & 69.9* & -- & -- & 95.6* & --
\\ 
\mbox{FedSurF-C-IPCW} & -- & 79.8* & -- & -- & \textbf{75.4}* & -- & -- & 70.2* & -- & -- & 69.9* & -- & -- & 95.6* & --
\\ 
\mbox{FedSurF-IBS} & -- & 79.8* & -- & -- & 75.1* & -- & -- & 70.3* & -- & -- & 70.2* & -- & -- & \textbf{95.7}* & --
\\ 
\mbox{FedSurF-AUC} & -- & \textbf{80.3}* & -- & -- & 75.0* & -- & -- & 70.5* & -- & -- & 70.7* & -- & -- & 95.6* & --
\\ 
\bottomrule
\end{tabularx}

\end{table*}

\myparagraph{Uniformly Split Data Across 10 Clients.} The performance metrics of survival models across five datasets -- WHAS500, GBSG2, METABRIC, NWTCO, and FLCHAIN -- are illustrated in Tables~\ref{tab:sim_cid_uni}, \ref{tab:sim_ibs_uni}, and \ref{tab:sim_auc_uni}. Label-skewed splitting~\cite{archetti2023heterogeneous} with $\alpha\rightarrow\infty$ was employed for data assignment to simulate federations with uniform data distribution. This analysis centers around federations comprising 10 clients. The metrics presented include the Concordance Index with IPCW weighting (C-Index-IPCW), Integrated Brier Score (IBS), and Cumulative AUC. The mean values across 20 runs are reported. Metrics for the \emph{Local}, \emph{Federated}, and \emph{Global} settings are given for each dataset. The Kruskal-Wallis test, followed by a pairwise Dunn's test at a significance level of 0.05, was conducted to assess statistical differences in the results. We focus on FedSurF-C for its efficient evaluation metric that does not necessitate IPCW weights. Any results not showing a statistically significant difference with the FedSurF-C performance are marked with an asterisk (*).

The results indicate that federated learning, as compared to local training, is advantageous for all clients on average from a performance perspective. In fact, all tables demonstrate superior performance in the \emph{Federated} setting compared to the \emph{Local} setting. Furthermore, the \emph{Federated} performance closely resembles the \emph{Global} performance, signifying that the performance gap between distributed and centralized learning is minimal while offering the added advantage of user privacy preservation in the \emph{Federated} setting.

Table~\ref{tab:sim_cid_uni}, which pertains to the C-Index IPCW, reveals that DeepSurv consistently performs well across datasets among the baseline models. FedSurF achieves comparable performance, particularly in WHAS500, where it surpasses all baselines. However, no clear winner emerges among the sampling techniques. In fact, sampling based on any of the proposed metrics yields better results on average than uniform sampling, but the difference is not statistically significant.

Regarding Table~\ref{tab:sim_ibs_uni} and IBS, survival forests do not outperform neural baselines. Nevertheless, their performance is comparable with no statistically significant difference, particularly in datasets with more samples (METABRIC, NWTCO, and FLCHAIN).

Lastly, Table~\ref{tab:sim_auc} presents the Cumulative AUC. Here, survival forests exhibit exceptional performance, where the best average AUC is achieved by one of the FedSurF variations, or within a non-statistically significant difference. The only difference is the NWTCO dataset, where DeepHit and N-MTLR exhibit better AUC than FedSurF models.

In summary, when data are uniformly split, FedSurF effectively enhances model performance compared to local models. The FedSurF variations consistently achieve robust performance across diverse evaluation metrics and datasets. However, any sampling strategy produces results close to the best. 

\myparagraph{Label-skewed Data Across 10 Clients.} Adopting the same experimental methodology, Tables~\ref{tab:sim_cid}, \ref{tab:sim_ibs}, and \ref{tab:sim_auc} illustrate performance metrics for survival models assessed in federations handling heterogeneous data. The data allocation was conducted utilizing label-skewed splitting\cite{archetti2023heterogeneous} with a parameter value of $\alpha=5$. This evaluation focuses on federations consisting of 10 clients.

In reference to Table~\ref{tab:sim_cid}, variations of the FedSurF algorithm demonstrate similar or superior concordance compared to the neural models, particularly in the context of smaller datasets (WHAS500 and GBSG2). The FLCHAIN dataset is the only exception where FedSurF variants did not perform at par with the top model, albeit the discrepancy was only by a small margin of a few percentage points.

Analyzing from the perspective of the IBS as presented in Table~\ref{tab:sim_ibs}, the FedSurF variations tend to rank towards the lower end of the spectrum. In the case of the NWTCO and FLCHAIN datasets, the N-MTLR model exhibits better IBS. However, for the remaining datasets, while FedSurF's average IBS is generally higher, the gap between it and the best-performing model is not statistically significant.

Lastly, Table~\ref{tab:sim_auc} showcases a promising trend in AUC, with FedSurF models either outperforming or matching the average of other models. NWTCO emerges as the sole exception, where N-MTLR outpaces all other alternatives.

From these results, it is evident that FedSurF variations perform comparably or even surpass neural baselines. With heterogeneously distributed data, FedSurF models typically exhibit better performance than neural models compared to when the data are uniformly split. This suggests that our algorithm is more resilient to federations comprising clients with different data distributions and dataset cardinalities.

Moreover, any FedSurF variation attains roughly the same performance with a slight disadvantage for FedSurF with uniform sampling. Consequently, we recommend employing FedSurF-C, as it is remarkably simple to compute without necessitating integration or IPCW weighting for evaluation. FedSurF++ can thus be considered a viable alternative to neural-based architectures for large-scale survival analysis, as it attains comparable performance with just a single model exchange round.

\begin{table*}[ht!]
\centering
\caption{Concordance Index with IPCW weighting (C-Index-IPCW)~\cite{uno2011c,robins1992recovery} for survival models evaluated on simulated federations ($K=10$, $\alpha=5$). Each C-Index-IPCW is scaled by a factor of 100 for better readability. We report the mean computed over 20 runs. The best results ($\uparrow$) are highlighted in bold. Values marked with * do not exhibit statistically significant differences with FedSurF-C according to Dunn's test with 0.05 significance.}
\label{tab:sim_cid}
    
\begin{tabularx}{\textwidth}{@{} l YYY YYY YYY YYY YYY}
\toprule
& \multicolumn{3}{c}{WHAS500} & \multicolumn{3}{c}{GBSG2} & \multicolumn{3}{c}{METABRIC} & \multicolumn{3}{c}{NWTCO} & \multicolumn{3}{c}{FLCHAIN}\\
\cmidrule(l){2-4}\cmidrule(l){5-7}\cmidrule(l){8-10}\cmidrule(l){11-13}\cmidrule(l){14-16}
Model & \emph{Loc.} & \emph{Fed.} & \emph{Glob.} & \emph{Loc.} & \emph{Fed.} & \emph{Glob.} & \emph{Loc.} & \emph{Fed.} & \emph{Glob.}& \emph{Loc.} & \emph{Fed.} & \emph{Glob.}& \emph{Loc.} & \emph{Fed.} & \emph{Glob.}\\ 
\midrule
\mbox{CoxPH} & 65.0 & 74.3 & 77.8 & 56.7 & 61.8 & 63.9 & 57.3 & 61.6 & 64.5 & 63.2 & 62.7 & 52.7 & 88.7 & 89.7 & 94.2
\\
\mbox{DeepSurv} & 68.8 & 77.3 & 73.4 & 58.0 & 65.1* & 65.7 & 60.6 & \textbf{64.2}* & 64.8 & 65.4 & 67.3* & 56.4 & 93.1 & 89.6* & 94.3
\\ 
\mbox{DeepHit} & 67.4 & 77.1 & 75.2 & 57.4 & 63.0 & 64.2 & 57.2 & 61.3 & 62.0 & 52.1 & 68.8* & 71.6 & 92.5 & 93.6 & 94.1
\\ 
\mbox{N-MTLR} & 68.5 & 76.6 & 74.6 & 58.2 & 64.8* & 63.7 & 60.0 & 63.8* & 64.5 & 64.9 & \textbf{70.3}* & 71.8 & 93.5 & \textbf{94.2} & 94.1
\\ 
\mbox{Nnet-Survival} & 65.2 & 74.6 & 76.1 & 55.1 & 61.4 & 63.5 & 50.1 & 58.6 & 62.7 & 45.8 & 66.1 & 70.3 & 87.6 & 93.7* & 94.2
\\ 
\mbox{PC-Hazard} & 66.0 & 75.1 & 75.6 & 55.1 & 61.1 & 63.6 & 50.3 & 58.4 & 62.8 & 45.8 & 66.1 & 70.3 & 86.7 & 93.8* & 94.2
\\ 
\midrule
\mbox{FedSurF} & 73.0 & 78.4* & 78.5 & 61.9 & 65.2* & 64.2 & 60.7 & 63.8* & 64.0 & 67.7 & 69.2* & 68.0 & 93.6 & 93.8* & 93.8
\\ 
\mbox{FedSurF-C} & -- & \textbf{79.3}* & -- & -- & 65.1* & -- & -- & 63.8* & -- & -- & 69.1* & -- & -- & 93.8* & --
\\ 
\mbox{FedSurF-C-IPCW} & -- & 79.1* & -- & -- & \textbf{65.4}* & -- & -- & 63.8* & -- & -- & 69.1* & -- & -- & 93.8* & --
\\ 
\mbox{FedSurF-IBS} & -- & 79.0* & -- & -- & \textbf{65.4}* & -- & -- & 63.8* & -- & -- & 69.1* & -- & -- & 93.8* & --
\\ 
\mbox{FedSurF-AUC} & -- & 79.1* & -- & -- & \textbf{65.4}* & -- & -- & 63.7* & -- & -- & 69.2* & -- & -- & 93.8* & --
\\ 
\bottomrule
\end{tabularx}
\end{table*}%
\begin{table*}[ht!]
\centering
\caption{Integrated Brier Score (IBS)~\cite{graf1999assessment} for survival models evaluated on simulated federations ($K=10$, $\alpha=5$). Each IBS is scaled by a factor of 100 for better readability. We report the mean computed over 20 runs. The best results ($\downarrow$) are highlighted in bold. Values marked with * do not exhibit statistically significant differences with FedSurF-C according to Dunn's test with 0.05 significance.}
\label{tab:sim_ibs}
    
\begin{tabularx}{\textwidth}{@{} l YYY YYY YYY YYY YYY}
\toprule
& \multicolumn{3}{c}{WHAS500} & \multicolumn{3}{c}{GBSG2} & \multicolumn{3}{c}{METABRIC} & \multicolumn{3}{c}{NWTCO} & \multicolumn{3}{c}{FLCHAIN}\\
\cmidrule(l){2-4}\cmidrule(l){5-7}\cmidrule(l){8-10}\cmidrule(l){11-13}\cmidrule(l){14-16}
Model & \emph{Loc.} & \emph{Fed.} & \emph{Glob.} & \emph{Loc.} & \emph{Fed.} & \emph{Glob.} & \emph{Loc.} & \emph{Fed.} & \emph{Glob.}& \emph{Loc.} & \emph{Fed.} & \emph{Glob.}& \emph{Loc.} & \emph{Fed.} & \emph{Glob.}\\ 
\midrule
\mbox{CoxPH} & 18.6 & \textbf{16.4}* & 14.8 & 19.1 & 17.8* & 16.8 & 17.4 & 16.4* & 15.8 & 11.4 & 12.5 & 11.9 & 8.5 & 7.7 & 4.1
\\
\mbox{DeepSurv} & 19.1 & 16.5* & 21.5 & 19.1 & \textbf{16.9}* & 16.8 & 17.1 & \textbf{16.0}* & 16.2 & 11.2 & 10.6* & 11.8 & 5.7 & 6.1* & 4.2
\\ 
\mbox{DeepHit} & 19.6 & 16.7* & 16.5 & 20.1 & 18.3 & 17.9 & 17.8 & 17.4 & 16.3 & 14.9 & 10.9* & 10.4 & 6.5 & 4.5* & 4.3
\\ 
\mbox{N-MTLR} & 19.4 & 17.4* & 20.2 & 19.4 & 17.1* & 18.0 & 17.3 & \textbf{16.0}* & 16.1 & 11.7 & \textbf{10.3} & 10.1 & 4.9 & \textbf{4.2} & 4.4
\\ 
\mbox{Nnet-Survival} & 22.1 & 20.6 & 16.9 & 21.7 & 18.7 & 17.2 & 22.5 & 23.3 & 16.4 & 16.2 & 10.9* & 10.1 & 7.9 & 4.5* & 4.2
\\ 
\mbox{PC-Hazard} & 22.1 & 18.7 & 17.2 & 21.5 & 18.7 & 17.2 & 22.4 & 18.8 & 16.5 & 16.0 & 10.9* & 10.1 & 8.1 & 4.5* & 4.2
\\ 
\midrule
\mbox{FedSurF} & 18.2 & 17.5* & 17.5 & 18.7 & 18.0* & 18.2 & 17.4 & 16.2* & 16.2 & 11.3 & 11.1* & 11.0 & 4.8 & 4.6 & 4.1
\\ 
\mbox{FedSurF-C} & -- & 17.0* & -- & -- & 17.8* & -- & -- & 16.2* & -- & -- & 11.0* & -- & -- & 4.4* & --
\\ 
\mbox{FedSurF-C-IPCW} & -- & 17.1* & -- & -- & 17.8* & -- & -- & 16.2* & -- & -- & 11.0* & -- & -- & 4.4* & --
\\ 
\mbox{FedSurF-IBS} & -- & 17.1* & -- & -- & 17.9* & -- & -- & 16.2* & -- & -- & 11.0* & -- & -- & 4.3* & --
\\ 
\mbox{FedSurF-AUC} & -- & 17.1* & -- & -- & 17.8* & -- & -- & 16.2* & -- & -- & 11.0* & -- & -- & 4.4* & --
\\ 
\bottomrule
\end{tabularx}
\end{table*}%
\begin{table*}[ht!]
\centering
\caption{Cumulative AUC~\cite{sksurv} for survival models evaluated on simulated federations ($K=10$, $\alpha=5$). Each Cumulative AUC is scaled by a factor of 100 for better readability. We report the mean computed over 20 runs. The best results ($\uparrow$) are highlighted in bold. Values marked with * do not exhibit statistically significant differences with FedSurF-C according to Dunn's test with 0.05 significance.}
\label{tab:sim_auc}
\begin{tabularx}{\textwidth}{@{} l YYY YYY YYY YYY YYY}
\toprule
& \multicolumn{3}{c}{WHAS500} & \multicolumn{3}{c}{GBSG2} & \multicolumn{3}{c}{METABRIC} & \multicolumn{3}{c}{NWTCO} & \multicolumn{3}{c}{FLCHAIN}\\
\cmidrule(l){2-4}\cmidrule(l){5-7}\cmidrule(l){8-10}\cmidrule(l){11-13}\cmidrule(l){14-16}
Model & \emph{Loc.} & \emph{Fed.} & \emph{Glob.} & \emph{Loc.} & \emph{Fed.} & \emph{Glob.} & \emph{Loc.} & \emph{Fed.} & \emph{Glob.}& \emph{Loc.} & \emph{Fed.} & \emph{Glob.}& \emph{Loc.} & \emph{Fed.} & \emph{Glob.}\\ 
\midrule
\mbox{CoxPH} & 65.7 & 75.8 & 79.4 & 61.2 & 70.3 & 74.6 & 59.5 & 65.4 & 68.9 & 64.1 & 64.4* & 53.2 & 90.3 & 91.5 & 95.6
\\
\mbox{DeepSurv} & 69.1 & 77.5* & 73.2 & 62.9 & 73.9* & 73.4 & 63.8 & 68.9* & 68.9 & 66.7 & 67.8* & 56.7 & 94.3 & 91.3* & 95.9
\\ 
\mbox{DeepHit} & 66.6 & 76.0 & 74.5 & 60.7 & 71.3 & 70.6 & 60.3 & 66.4 & 69.7 & 52.3 & 72.9 & 74.5 & 93.7 & 95.3 & 95.4
\\ 
\mbox{N-MTLR} & 68.4 & 76.4 & 73.8 & 62.6 & 72.7* & 72.0 & 64.1 & \textbf{71.0}* & 72.3 & 67.8 & \textbf{73.5} & 74.6 & 94.6 & \textbf{95.6}* & 95.8
\\ 
\mbox{Nnet-Survival} & 62.9 & 72.5 & 75.9 & 57.8 & 68.6 & 72.7 & 48.7 & 60.3 & 69.4 & 37.7 & 67.5* & 72.4 & 88.6 & 95.2 & 95.8
\\ 
\mbox{PC-Hazard} & 63.1 & 72.4 & 74.7 & 57.9 & 68.8 & 72.9 & 48.5 & 61.1 & 69.3 & 37.9 & 67.2* & 71.9 & 87.2 & 94.8 & 95.3
\\ 
\midrule
\mbox{FedSurF} & 73.5 & 78.8* & 79.9 & 69.0 & 74.9* & 73.6 & 64.9 & 70.1* & 70.9 & 68.3 & 69.9* & 68.5 & 94.8 & 95.5* & 96.1
\\ 
\mbox{FedSurF-C} & -- & \textbf{79.8}* & -- & -- & 74.8* & -- & -- & 70.2* & -- & -- & 69.8* & -- & -- & \textbf{95.6}* & --
\\ 
\mbox{FedSurF-C-IPCW} & -- & 79.3* & -- & -- & \textbf{75.2}* & -- & -- & 70.2* & -- & -- & 69.9* & -- & -- & 95.5* & --
\\ 
\mbox{FedSurF-IBS} & -- & 79.4* & -- & -- & 75.1* & -- & -- & 70.2* & -- & -- & 70.3* & -- & -- & \textbf{95.6}* & --
\\ 
\mbox{FedSurF-AUC} & -- & \textbf{79.8}* & -- & -- & \textbf{75.2}* & -- & -- & 70.3* & -- & -- & 70.8* & -- & -- & 95.5* & --
\\ 
\bottomrule
\end{tabularx}
\end{table*}

\begin{figure*}[t!]
    \centering
    \includegraphics[width=\textwidth]{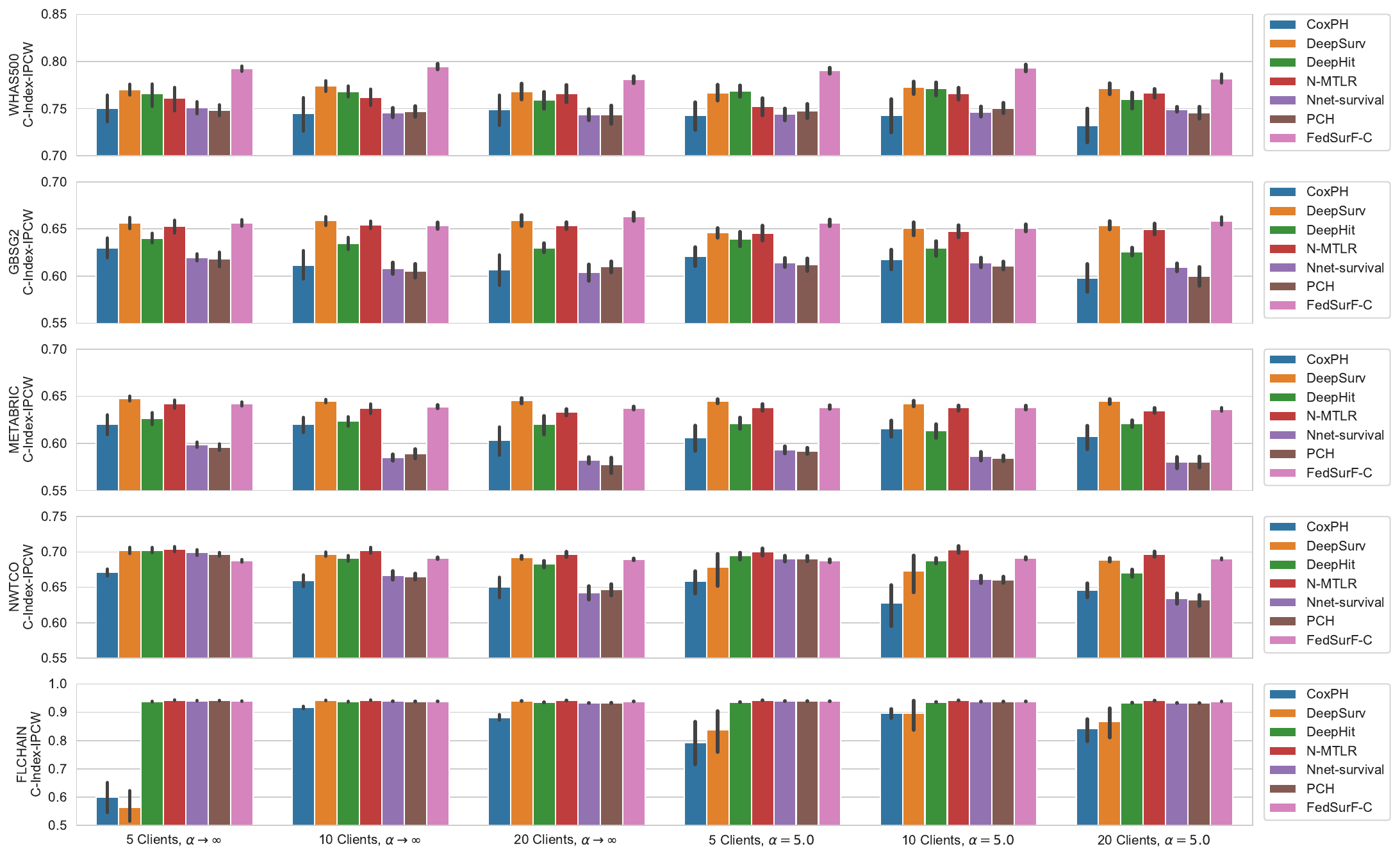}
    \caption{C-Index-IPCW metrics for several federation configurations. Specifically, the number of clients $K$ can be 5, 10, or 20 and the label splitting parameter $\alpha$ either tends to infinity (federation with uniformly split data) or equals 5 (federation with heterogeneous data distributions). Each row corresponds to a survival dataset and each bar to a different survival model. Results are averaged over 20 runs.}
    \label{fig:diff_k}
\end{figure*}

\myparagraph{Federations with Varied Numbers of Clients.} To conclude our analysis of simulated federations, we evaluated federations that consisted of a varying number of clients. Specifically, we assessed federations with 5, 10, and 20 clients. The results are summarized in Figure~\ref{fig:diff_k}. For simplicity, we chose to present only the C-Index-IPCW metrics and focused on the FedSurF-C variation among the FedSurF models.

These results highlight the robustness of FedSurF-C across diverse client configurations. Notably, the performance of FedSurF-C remains consistent regardless of whether the number of clients increased or decreased. This consistency is not matched by neural baselines. For instance, the performance of Nnet-survival and PCH tends to decline as the number of clients increases. Conversely, proportional hazard models (CoxPH and DeepSurv) do not exhibit a performance trend that is proportional to the number of clients. In fact, their results do markedly vary when the number of clients is modified, displaying high variance and thus proving less reliable than the alternatives.

To summarize, although FedSurF-C may not outperform neural models in all configurations, it does display the most consistent concordance when varying the number of clients.

\subsection{Experiments on Real-World Federations}
\label{sec:real_world_federations}

This section covers the experiments related to real-world heterogeneous datasets, Lombardy Heart Failure~\cite{Mazzali2016MethodologicalIO} (Section~\ref{sec:lombardy}) and Fed-TCGA-BRCA~\cite{terrail2022flamby} (Section~\ref{sec:flamby}).

Experiments on real federated data follow the same procedures and hyperparameters as in simulated federations. Therefore, neural networks have the same structure and training follows the same number of local epochs and rounds of federated averaging. 30\% of the local data sets are selected for validation. The same metrics described in Section~\ref{sec:sf_evaluation} are used to evaluate the methods. The only difference is that the data are already distributed across multiple clients, so label-skewed splitting~\cite{archetti2023heterogeneous} is not applied to create federations.

\subsubsection{The Lombardy Heart Failure Dataset}
\label{sec:lombardy}

The Lombardy Heart Failure administrative dataset~\cite{Mazzali2016MethodologicalIO} was derived from the HFData research project (RF-2009-1483329), which aimed to examine heart failure cases in Lombardy between 2000 and 2012. Lombardy, one of Italy's largest and most populous regions, has a population of approximately 10 million individuals, accounting for 16.5\% of the nation's total population. The dataset was provided by the Regione Lombardia -- Healthcare Division and pertains to non-pediatric residents who were hospitalized for heart failure between January 2006 and December 2012. Hospital discharge charts (HDC) were employed to gather information about patients' hospitalization, including the discharge date, length of stay, and comorbidity conditions. Additionally, information about pharmaceutical purchases was obtained from the Anatomical Therapeutic Chemical (ATC) codes. The dataset offers a detailed view of patients' clinical histories of hospitalizations. A description of the preliminary preprocessing and collection can be found in~\cite{Mazzali2016MethodologicalIO}.

The initial dataset includes 339,690 samples with 48 features, including several hospitalizations and pharmaceutical prescriptions per patient. For our case study, we focused solely on hospitalizations, reducing the original data to 22,418 samples. To ensure that each patient had a follow-up of at least 5 years and to create a comparable cohort for our algorithms, we eliminated new hospitalizations between 2008 and 2012. This left us with data from 2006 to 2007, with updated time labels to match each patient's actual outcome. Administrative censoring was employed for patients who survived until the end of 2012. Additionally, we removed features related to pharmacological prescriptions, resulting in a dataset with 32 covariates.

\begin{figure}[t!]
    \centering
    \includegraphics[width=\linewidth]{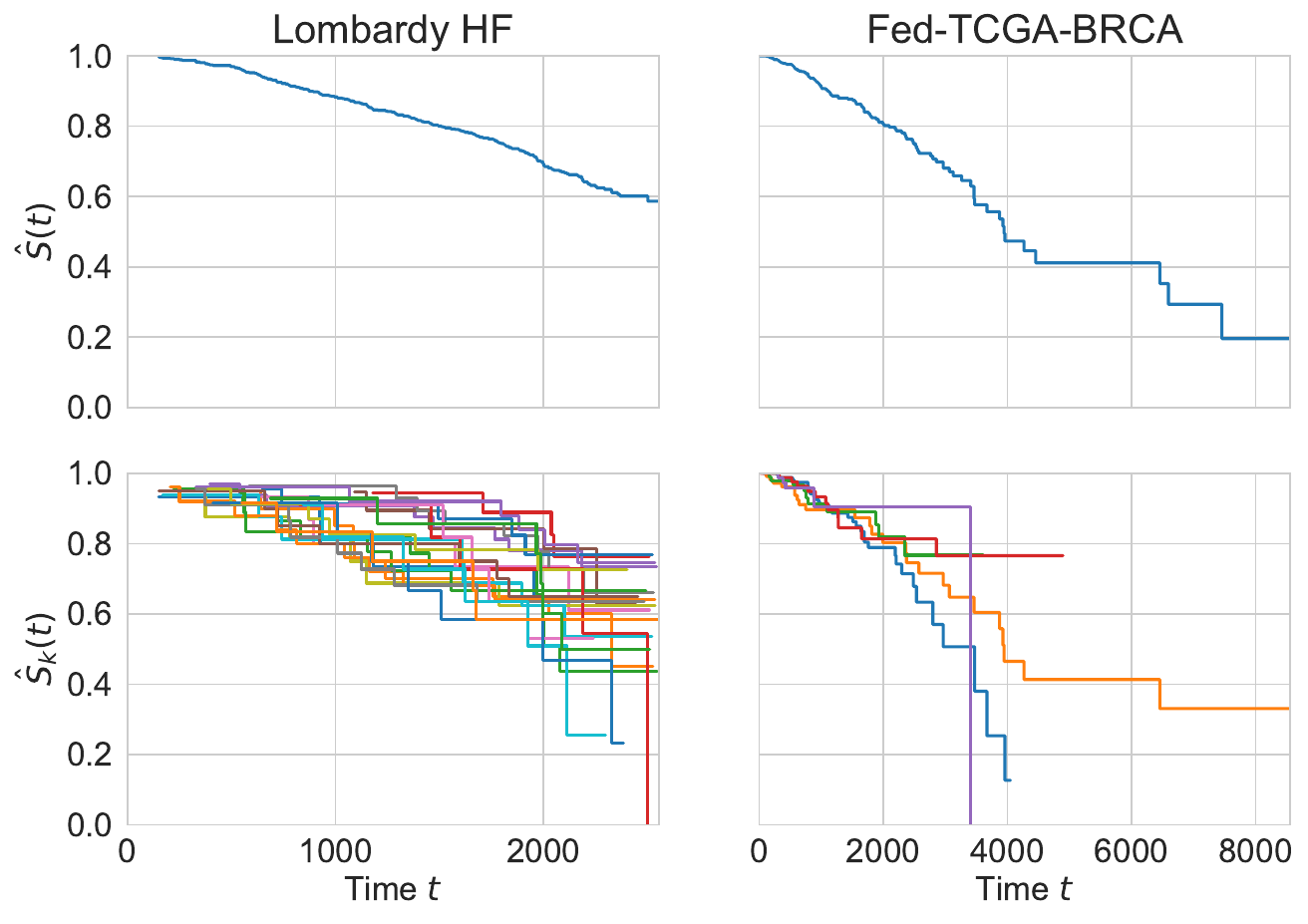}
    \caption{Kaplan-Meier estimators $\hat{S}(t)$ for datasets of real-world federations. The first row shows KM estimators for the entire dataset, while the second row depicts a KM curve for each client $k$.}
    \label{fig:real_km}
\end{figure}

Finally, the dataset was split into the federation of medical structures where hospitalization occurred. We excluded medical structures with fewer than 10 events or fewer than 20 total samples in their local data, leaving us with 895 samples distributed across 23 clients. Figure~\ref{fig:real_km} shows the Kaplan-Meier estimators of the entire dataset and the ones related to each client. From these plots, client distributions exhibit significantly different patterns.

\subsubsection{The Fed-TCGA-BRCA Dataset}
\label{sec:flamby}

The Fed-TCGA-BRCA survival dataset~\cite{terrail2022flamby} is a federated dataset for survival analysis based on clinical data from The Cancer Genome Atlas (TCGA) project. TCGA is a large-scale initiative that aims to characterize the genomic changes in various types of cancer. BRCA stands for breast invasive carcinoma, which is one of the cancer types studied by TCGA. The Fed-TCGA-BRCA survival dataset contains survival outcomes for 1088 patients with BRCA and 38 binary features for each patient. The dataset is distributed among six regions (Northeast, South, West, Midwest, Europe, and Canada) based on the tissue sort site (TSS) of each patient. Each of these regions contains 248, 156, 164, 129, 129, and 40 samples respectively. In particular, among the 40 samples from Canada, only a single entry exhibits an event. If the Canada dataset would be split into training and validation, one of the splits would contain no events. For that split, it would not be possible to evaluate the concordance index, as there would be no comparable pairs of subjects. For this reason, we excluded the client corresponding to Canada for training and retained the other five regions. Figure~\ref{fig:real_km} shows the Kaplan-Meier estimators of the entire dataset and the ones related to each client.

\subsubsection{Results on Lombardy HF and Fed-TCGA-BRCA}
\label{sec:flamby_results}

\begin{table*}[ht!]
\centering
\caption{Concordance Index with IPCW weighting (C-Index-IPCW)~\cite{uno2011c,robins1992recovery} for survival models evaluated on real-world federations. Each C-Index-IPCW is scaled by a factor of 100 for better readability. We report the mean and the standard deviation computed over 20 runs. The best results ($\uparrow$) are highlighted in bold. Values marked with * do not exhibit statistically significant differences with FedSurF-C according to Dunn's test with 0.05 significance.}
\label{tab:real_cid}
    
\begin{tabularx}{\textwidth}{@{} l YYY YYY}
\toprule
& \multicolumn{3}{c}{Lombardy Heart Failure} & \multicolumn{3}{c}{Fed-TCGA-BRCA} \\
\cmidrule(l){2-4}\cmidrule(l){5-7}
Model & \emph{Local} & \emph{Federated} & \emph{Global} & \emph{Local} & \emph{Federated} & \emph{Global} \\ 
\midrule
\mbox{CoxPH} & 58.4 $\pm$ 1.4 & 69.2 $\pm$ 2.4 & 71.4 $\pm$ 0.4 & 60.3 $\pm$ 5.5 & 75.4 $\pm$ 4.6* & 77.0 $\pm$ 0.9
\\
\mbox{DeepSurv} & 59.9 $\pm$ 1.9 & 71.7 $\pm$ 0.9 & 70.6 $\pm$ 0.8 & 61.8 $\pm$ 4.4 & \textbf{78.9 $\pm$ 2.1}* & 72.3 $\pm$ 2.5
\\ 
\mbox{DeepHit} & 57.6 $\pm$ 1.4 & 69.1 $\pm$ 2.0 & 71.4 $\pm$ 0.9 & 59.3 $\pm$ 4.2 & 76.4 $\pm$ 5.7* & 79.9 $\pm$ 2.1
\\ 
\mbox{N-MTLR} & 57.7 $\pm$ 1.9 & 70.9 $\pm$ 1.2 & 69.6 $\pm$ 0.8 & 60.6 $\pm$ 4.5 & 77.7 $\pm$ 3.4* & 75.0 $\pm$ 3.2
\\ 
\mbox{Nnet-Survival} & 50.9 $\pm$ 1.2 & 69.2 $\pm$ 1.6 & 71.7 $\pm$ 0.5 & 55.9 $\pm$ 3.4 & 71.4 $\pm$ 3.0 & 77.4 $\pm$ 2.4
\\ 
\mbox{PC-Hazard} & 50.8 $\pm$ 1.2 & 69.2 $\pm$ 1.3 & 71.4 $\pm$ 0.4 & 56.9 $\pm$ 2.8 & 68.5 $\pm$ 7.4 & 76.3 $\pm$ 2.4
\\ 
\midrule
\mbox{FedSurF} & 61.7 $\pm$ 0.9 & 73.6 $\pm$ 0.9* & 72.7 $\pm$ 0.1 & 66.7 $\pm$ 3.1 & 76.3 $\pm$ 2.3* & 72.3 $\pm$ 0.7
\\ 
\mbox{FedSurF-C} & -- & 73.5 $\pm$ 0.7* & -- & -- & 77.2 $\pm$ 2.1* & --
\\ 
\mbox{FedSurF-C-IPCW} & -- & 73.6 $\pm$ 0.6* & -- & -- & 76.7 $\pm$ 2.1* & --
\\ 
\mbox{FedSurF-IBS} & -- & \textbf{73.7 $\pm$ 0.8}* & -- & -- & 76.9 $\pm$ 1.8* & --
\\ 
\mbox{FedSurF-AUC} & -- & 73.6 $\pm$ 0.5* & -- & -- & 77.1 $\pm$ 1.9* & --
\\ 
\bottomrule
\end{tabularx}

\end{table*}

\begin{table*}[ht!]
\centering
\caption{Integrated Brier Score (IBS)~\cite{graf1999assessment} for survival models evaluated on real-world federations. Each IBS is scaled by a factor of 100 for better readability. We report the mean and the standard deviation computed over 20 runs. The best results ($\downarrow$) are highlighted in bold. Values marked with * do not exhibit statistically significant differences with FedSurF-C according to Dunn's test with 0.05 significance.}
\label{tab:real_ibs}
    
\begin{tabularx}{\textwidth}{@{} l YYY YYY}
\toprule
& \multicolumn{3}{c}{Lombardy Heart Failure} & \multicolumn{3}{c}{Fed-TCGA-BRCA} \\
\cmidrule(l){2-4}\cmidrule(l){5-7}
Model & \emph{Local} & \emph{Federated} & \emph{Global} & \emph{Local} & \emph{Federated} & \emph{Global} \\ 
\midrule
\mbox{CoxPH} & 13.6 $\pm$ 0.1 & 13.0 $\pm$ 0.3* & 12.4 $\pm$ 0.1 & 28.2 $\pm$ 1.9 & 24.8 $\pm$ 2.1* & 24.7 $\pm$ 0.4
\\
\mbox{DeepSurv} & 14.0 $\pm$ 0.3 & \textbf{12.3 $\pm$ 0.3} & 13.0 $\pm$ 0.2 & 28.8 $\pm$ 1.2 & 25.7 $\pm$ 1.6 & 32.3 $\pm$ 4.3
\\ 
\mbox{DeepHit} & 17.5 $\pm$ 0.3 & 14.3 $\pm$ 3.6* & 12.5 $\pm$ 0.1 & 31.2 $\pm$ 0.4 & 28.2 $\pm$ 2.4 & 27.3 $\pm$ 2.2
\\ 
\mbox{N-MTLR} & 15.3 $\pm$ 0.3 & 12.5 $\pm$ 0.3 & 13.2 $\pm$ 0.3 & 29.5 $\pm$ 2.1 & 26.6 $\pm$ 4.1 & 36.2 $\pm$ 5.6
\\ 
\mbox{Nnet-Survival} & 20.2 $\pm$ 0.4 & 12.8 $\pm$ 0.4* & 12.5 $\pm$ 0.1 & 47.3 $\pm$ 1.1 & 37.0 $\pm$ 2.8 & 26.5 $\pm$ 1.2
\\ 
\mbox{PC-Hazard} & 20.0 $\pm$ 0.5 & 12.8 $\pm$ 0.3* & 12.6 $\pm$ 0.1 & 46.7 $\pm$ 1.2 & 43.7 $\pm$ 14.9 & 26.3 $\pm$ 0.8
\\ 
\midrule
\mbox{FedSurF} & 14.2 $\pm$ 0.1 & 13.3 $\pm$ 0.1* & 12.1 $\pm$ 0.0 & 26.5 $\pm$ 0.7 & 23.2 $\pm$ 0.4* & 24.2 $\pm$ 0.3
\\ 
\mbox{FedSurF-C} & -- & 13.1 $\pm$ 0.1* & -- & -- & \textbf{22.9 $\pm$ 0.3}* & --
\\ 
\mbox{FedSurF-C-IPCW} & -- & 13.1 $\pm$ 0.1* & -- & -- & 23.1 $\pm$ 0.4* & --
\\ 
\mbox{FedSurF-IBS} & -- & 13.2 $\pm$ 0.0* & -- & -- & 23.1 $\pm$ 0.3* & --
\\ 
\mbox{FedSurF-AUC} & -- & 13.1 $\pm$ 0.0* & -- & -- & 23.0 $\pm$ 0.3* & --
\\ 
\bottomrule
\end{tabularx}

\end{table*}

\begin{table*}[ht!]
\centering
\caption{Cumulative AUC~\cite{sksurv} for survival models evaluated on real-world federations. Each Cumulative AUC is scaled by a factor of 100 for better readability. We report the mean and the standard deviation computed over 20 runs. The best results ($\uparrow$) are highlighted in bold. Values marked with * do not exhibit statistically significant differences with FedSurF-C according to Dunn's test with 0.05 significance.}
\label{tab:real_auc}
    
\begin{tabularx}{\textwidth}{@{} l YYY YYY}
\toprule
& \multicolumn{3}{c}{Lombardy Heart Failure} & \multicolumn{3}{c}{Fed-TCGA-BRCA} \\
\cmidrule(l){2-4}\cmidrule(l){5-7}
Model & \emph{Local} & \emph{Federated} & \emph{Global} & \emph{Local} & \emph{Federated} & \emph{Global} \\ 
\midrule
\mbox{CoxPH} & 58.6 $\pm$ 1.4 & 70.5 $\pm$ 2.7 & 73.6 $\pm$ 0.3 & 62.7 $\pm$ 6.1 & 79.7 $\pm$ 3.6 & 77.6 $\pm$ 0.5
\\
\mbox{DeepSurv} & 60.5 $\pm$ 2.0 & 73.5 $\pm$ 1.1* & 74.2 $\pm$ 1.1 & 63.4 $\pm$ 4.9 & \textbf{80.7 $\pm$ 2.0} & 71.0 $\pm$ 4.3
\\ 
\mbox{DeepHit} & 57.6 $\pm$ 1.2 & 70.4 $\pm$ 2.4 & 72.4 $\pm$ 1.1 & 59.0 $\pm$ 5.0 & 75.0 $\pm$ 5.5* & 77.2 $\pm$ 2.1
\\ 
\mbox{N-MTLR} & 57.4 $\pm$ 2.2 & 71.9 $\pm$ 1.1 & 69.9 $\pm$ 1.3 & 61.3 $\pm$ 5.8 & 77.2 $\pm$ 5.0* & 72.7 $\pm$ 4.0
\\ 
\mbox{Nnet-Survival} & 51.2 $\pm$ 1.3 & 70.3 $\pm$ 1.9 & 73.5 $\pm$ 0.8 & 55.7 $\pm$ 3.2 & 71.1 $\pm$ 2.4* & 75.7 $\pm$ 3.0
\\ 
\mbox{PC-Hazard} & 51.1 $\pm$ 1.2 & 70.2 $\pm$ 1.9 & 72.9 $\pm$ 0.7 & 55.3 $\pm$ 2.3 & 67.0 $\pm$ 8.1* & 76.0 $\pm$ 2.9
\\ 
\midrule
\mbox{FedSurF} & 60.3 $\pm$ 0.9 & \textbf{74.9 $\pm$ 1.3}* & 73.4 $\pm$ 0.2 & 64.5 $\pm$ 3.0 & 72.9 $\pm$ 2.1* & 72.1 $\pm$ 0.5
\\ 
\mbox{FedSurF-C} & -- & 74.8 $\pm$ 1.0* & -- & -- & 73.8 $\pm$ 1.6* & --
\\ 
\mbox{FedSurF-C-IPCW} & -- & \textbf{74.9 $\pm$ 1.1}* & -- & -- & 73.0 $\pm$ 1.9* & --
\\ 
\mbox{FedSurF-IBS} & -- & 74.8 $\pm$ 1.2* & -- & -- & 73.9 $\pm$ 1.7* & --
\\ 
\mbox{FedSurF-AUC} & -- & \textbf{74.9 $\pm$ 1.0}* & -- & -- & 73.7 $\pm$ 1.4* & --
\\ 
\bottomrule
\end{tabularx}

\end{table*}

Table~\ref{tab:real_cid}, Table~\ref{tab:real_ibs}, and Table~\ref{tab:real_auc} display performance metrics for neural and ensemble-based survival models evaluated on two real-world federations: Lombardy Heart Failure and Fed-TCGA-BRCA. The metrics include Concordance Index with IPCW weighting (C-Index-IPCW), Integrated Brier Score (IBS), and Cumulative AUC. The reported results represent the mean and standard deviation computed over 20 runs. The Kruskal-Wallis and Dunn's tests with a significance level of 0.05 are conducted to assess statistical differences in the results. Any results not showing a statistically significant difference with FedSurF-C are marked with an asterisk (*).

Experiments on real-world federations corroborate the trend that participating in a federation yields superior results compared to training solely on local data. In fact, for most models, the \emph{Federated} results surpass the \emph{Local} results, occasionally even slightly exceeding the performance of \emph{Global} models.

Table~\ref{tab:real_cid} (C-Index-IPCW) indicates that FedSurF-IBS achieves the best performance for Lombardy HF, while DeepSurv exceeds the other models for Fed-TCGA-BRCA. However, for the latter case, no statistically significant difference exists between DeepSurv and FedSurF-C. Consequently, FedSurF variations have a strong discriminative power on real-world data, accurately identifying patients at risk in most cases. Again, the specific metric for sampling trees does not significantly impact the final outcome.

Table~\ref{tab:real_ibs} (IBS) identifies DeepSurv as the most calibrated model for Lombardy HF, demonstrating its exceptional performance as a survival model when communication constraints are not a concern. In contrast, FedSurF-C exhibits better results by a noticeable margin in the Fed-TCGA-BRCA dataset compared to all neural alternatives.

Finally, Table~\ref{tab:real_auc} (Cumulative AUC) showcases the best model performance in terms of discriminative ability. Concerning Fed-TCGA-BRCA, DeepSurv outperforms any other model by a fair margin. Instead, for Lombardy HF, the FedSurF variations outperform neural models, with the only exception of DeepSurv, where the performance difference is not statistically significant.

In summary, FedSurF++ proves to be a valuable alternative to neural-network-based models concerning the most common survival metrics, C-Index-IPCW, IBS, and Cumulative AUC. DeepSurv is a robust alternative from a performance standpoint, but its training procedure needs iterative averaging of model parameters, resulting in substantial bandwidth usage. FedSurF, conversely, may occasionally exhibit lower -- yet still comparable -- performance metrics while requiring only a single tree exchange round to generate the final model. Regarding the sampling strategy, experiments on real-world data confirm that the metrics considered during sampling do not affect the final results to a significant degree. Therefore, FedSurF-C is the most straightforward choice, relying solely on the local evaluation of the concordance index.

\section{Conclusion}
\label{sec:conclusion}

In this paper, we presented an extension of the Federated Survival Forest (FedSurF++) algorithm, which applies Random Survival Forests (RSFs) to a federated learning setting. The FedSurF++ algorithm builds upon the original FedSurF by introducing new tree sampling strategies, including concordance index, IPCW concordance index, integrated Brier score, and cumulative AUC. These strategies enable the selection of the best-performing trees from local RSF models, consequently improving the performance of the global RSF model.

Our experimental results on synthetic and real-world clinical trial datasets, covering heart failure and breast cancer genomics, demonstrate the effectiveness of FedSurF++ in various federations. The algorithm outperforms local models and achieves performance metrics comparable to global models, showcasing its robustness across diverse evaluation metrics and datasets. In particular, the FedSurF++ family consistently attains strong performance across different evaluation metrics and datasets. Moreover, our findings reveal that any sampling strategy, except for uniform sampling, yields results close to the best. 

While DeepSurv exhibits strong performance, its training procedure requires iterative averaging of model parameters, leading to heavy bandwidth usage. In contrast, FedSurF++ demands only a single tree exchange round to produce the final model, minimizing the communication overhead. In conclusion, the FedSurF++ algorithm proves to be a valuable alternative to neural-network-based models for large-scale survival analysis on confidential clinical data as it achieves comparable performance while preserving data privacy and offering an efficient solution in terms of communication.

\section*{Acknowledgment}
This project has been supported by AI-SPRINT: AI in Secure Privacy-pReserving computINg conTinuum (European Union H2020 grant agreement No. 101016577) and FAIR: Future Artificial Intelligence Research (NextGenerationEU, PNRR-PE-AI scheme, M4C2, investment 1.3, line on Artificial Intelligence).

\bibliographystyle{elsarticle-num}
\bibliography{references}

\end{document}